\documentclass[letterpaper, 10 pt, conference]{ieeeconf}  

\IEEEoverridecommandlockouts                              

\title{\LARGE \bf
Fauna Sprout: \\A lightweight, approachable, developer-ready humanoid robot 
}

\author{
Fauna Robotics Team
}

\usepackage{xcolor}   
\usepackage{graphicx}
\usepackage{siunitx}

\usepackage{multicol}
\usepackage{placeins}
\usepackage{booktabs}
\usepackage{multirow}

\usepackage{amsmath}
\usepackage{amssymb}  
\usepackage{mathtools}
\usepackage{bbm}
\usepackage{bm}
\usepackage{caption}
\usepackage[
  colorlinks=true,
  linkcolor=black,
  citecolor=black,
  urlcolor=blue
]{hyperref}

\begin{document}

\twocolumn[{%
\renewcommand\twocolumn[1][]{#1}%
\maketitle
\begin{center}
    \centering
    \captionsetup{type=figure}
    \includegraphics[width=\linewidth]{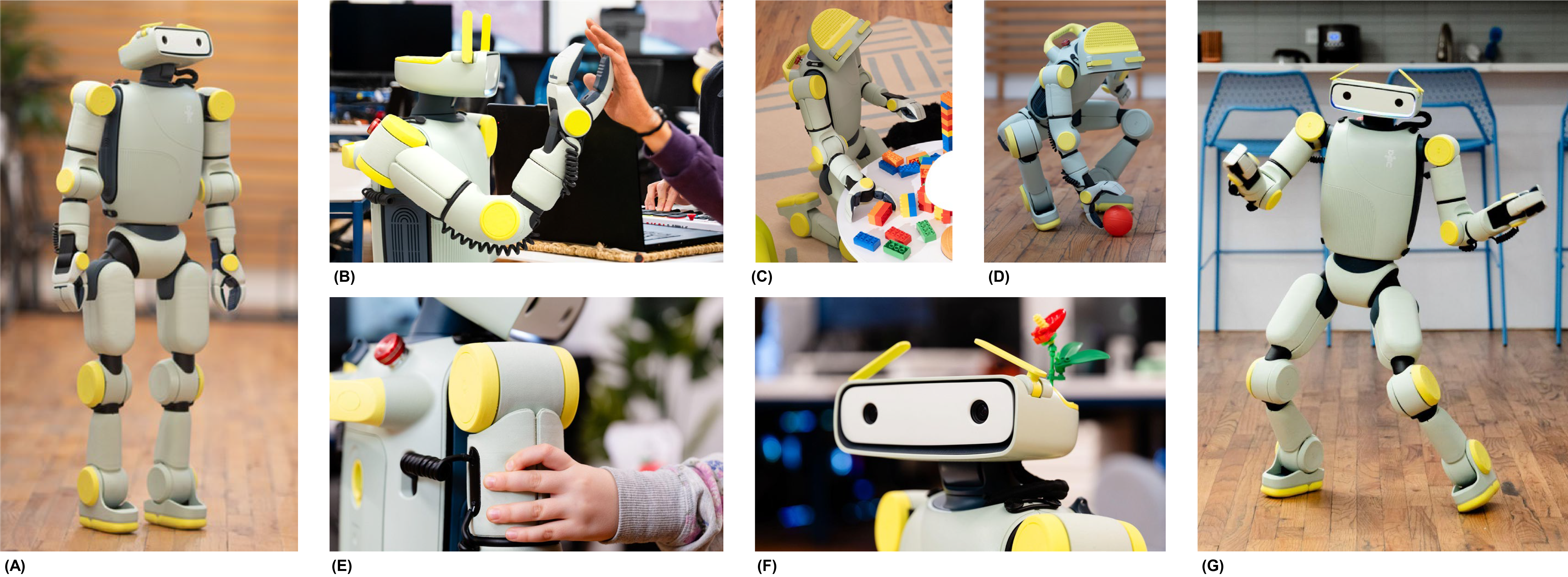}
    \caption{\textbf{Robot in action.} (A) Standing and looking up towards a person (B) performing closed-loop high-five interaction with a person (C) kneeling at a coffee table manipulating interlocking toy bricks (D) reaching the ground to pick up an object (E) being squeezed by a child's hand (F) posing with a toy flower attached to its head (G) dancing with expressive lights and eyebrow movements.}
    \label{fig:teaser}
\end{center}%
}]

\thispagestyle{empty}
\pagestyle{plain}

\begingroup
\renewcommand{\thefootnote}{}

\footnotetext{All authors are affiliated with Fauna Robotics, NYC, USA.}
\endgroup
        
\begin{abstract}
Recent advances in learned control, large-scale simulation, and generative models have accelerated progress toward general-purpose robotic controllers, yet the field still lacks platforms suitable for safe, expressive, long-term deployment in human environments. Most existing humanoids are either closed industrial systems or academic prototypes that are difficult to deploy and operate around people, limiting progress in robotics. We introduce Sprout, a developer platform designed to address these limitations through an emphasis on safety, expressivity, and developer accessibility. Sprout adopts a lightweight form factor with compliant control, limited joint torques, and soft exteriors to support safe operation in shared human spaces. The platform integrates whole-body control, manipulation with integrated grippers, and virtual-reality-based teleoperation within a unified hardware-software stack. An expressive head further enables social interaction---a domain that remains underexplored on most utilitarian humanoids. By lowering physical and technical barriers to deployment, Sprout expands access to capable humanoid platforms and provides a practical basis for developing embodied intelligence in real human environments. See \href{https://faunarobotics.com/}{faunarobotics.com} for videos.
\end{abstract}

\section{Introduction}

Progress in simulation, learned control, and generative AI has renewed interest in building humanoid robots capable of operating in everyday human environments. Despite these advances, researchers and developers face limited access to platforms that can be deployed safely and reliably around people. This absence is not merely an inconvenience, but a gap in the physical infrastructure required to develop embodied intelligence in human settings.

The most transformative computing technologies---personal computers, smartphones, and virtual reality---matured only after accessible, general-purpose platforms enabled broad participation and experimentation. Robotics has not yet reached this inflection point. The lack of accessible platforms inhibits the cadence of development, while safety constraints push applications toward short-horizon behavior in sanitized settings rather than sustained interaction in human environments. Participation by educators, designers, and creative technologists---groups whose practices emphasize exploration, iteration, and situated use---is further limited by the need for deep expertise in low-level control, calibration, and system maintenance. 

We introduce Sprout, a humanoid platform explicitly designed for safe, capable, and expressive operation in human environments. Standing \SI{1.07}{m} tall and weighing \SI{22.7}{kg}, Sprout has limited kinetic energy and correspondingly reduced potential impact forces that substantially improve its safety profile around people~\cite{Brooks2025HumanoidsDexterity}. Sprout complements this choice of scale with mechanical and control features intended to mitigate risk during close-proximity operation, including soft exterior panels, minimized pinch points, and backdrivable motors with conservative joint torque limits. At the control level, whole-body behaviors are executed through compliant controllers that minimize contact forces and reduce the likelihood of harm during incidental contact. Together, these design choices aim to support routine operation in shared human spaces rather than isolated or tightly controlled settings.

Sprout also serves as a capable and ergonomic platform for general robotics research, development, and application prototyping. The platform features a rear handle for manual maneuvering, a swappable battery for convenient day-to-day operation, and integrated, durable grippers designed to grasp across a wide range of everyday objects. The platform's modular hardware-software architecture provides core system services supporting different levels of abstraction. Low-level motor control development is served by APIs for sensing, actuation, and logging. Reusable whole-body controllers that handle locomotion and postural transitions enable development of high-level planning and reasoning. Integrated teleoperation, mapping, and navigation capabilities further support development of manipulation and autonomy. 

Sprout's character-like appearance is central to its purpose---to enable new modes of interaction that rely on social connection and physical presence. An articulated neck enables controllable gaze, while actuated eyebrows and an LED array support other interaction-relevant cues. In sum, Sprout lowers the barrier to creating meaningful robot interactions and broadens the set of people who can experiment with embodied intelligence.

\section{Related Work}

\subsection{Humanoid Platforms}

The landscape of humanoid robot development has expanded rapidly in recent years, spanning academic research, corporate R\&D, and a growing number of commercial startups. Yet despite this surge of activity, accessible platforms that are suitable for capable, safe, and expressive interaction around people remain limited, particularly within the United States. Existing humanoid systems tend to cluster into two categories: open academic projects that demonstrate impressive technical capabilities but limited manufacturability, and proprietary industrial efforts that are usually unavailable to independent developers.

Among open-source and academic systems, several recent efforts have advanced the state of humanoid design and control. Toddlerbot~\cite{shi2025toddlerbot} and earlier popular academic platforms, such as OP3~\cite{RobotisOP3_eManual}, illustrate the potential of open architectures and smaller-scale designs for educational and research use. Other impactful platforms include the Berkeley Humanoid~\cite{liao2024berkeley}, a mid-size bipedal locomotion platform, and ARTEMIS~\cite{zhu2025artemis}, a full-size bipedal platform designed to compete in robot soccer that demonstrated dynamic locomotion abilities. Collectively, these efforts highlight growing academic interest in open, replicable humanoid systems; however, few are designed for safe, long-term operation in close proximity to people, nor do they enable use by non-specialist developers.

Industrial and corporate efforts have generally focused on large, full-scale humanoids oriented toward logistics, manufacturing, or demonstration purposes (with the notable exception of entertainment-focused R\&D efforts, such as those by Disney Research~\cite{Grandia2024, mueller2025olaf}). Industrial examples include platforms from Boston Dynamics, Tesla, Agility, Figure, and XPENG. Fewer companies presently frame themselves as home-oriented (e.g., 1X Neo). In contrast, a new generation of Chinese startups has emphasized rapid manufacturability and cost-effective replication. Companies such as Unitree, Fourier Intelligence, LimX, Booster, and AgiBot have made significant strides in distributing humanoids to universities and developers worldwide. Of these, Unitree (G1, R1) and Booster (T1, K1) seem to have achieved the broadest uptake among U.S.-based students and researchers, owing to their relatively compact size, hardware reliability, and lower cost.

Morphology represents another important design consideration. While some modern systems employ legged locomotion, others adopt wheeled or hybrid “semi-humanoid” forms that trade the nimbleness and small footprint of feet for simplicity (e.g., Reflex Robotics and Sunday Robotics). Another critical differentiation lies in head design: most industrial humanoids omit expressive heads altogether for featureless glossy surfaces. Other expressive robots such as those by Anki, Enchanted Tools, and the Embodied Moxie display expression on digital screens. In contrast, we use non-screen show elements (coordinated lights and eyebrows) to emphasize the physicality of robotic embodiment and support non-verbal expressions, an approach more akin to Reachy Mini.

\subsection{Control System Architectures}

In recent years, the development of whole-body control for humanoid robots has been fundamentally reshaped by GPU-accelerated physics simulation. Modern frameworks such as NVIDIA's IsaacLab~\cite{mittal2025isaac} and Newton, MuJoCo~\cite{todorov2012mujoco} and its associated backends MuJoCo Warp and Mujoco XLA, and Genesis~\cite{Genesis} now support high-throughput rollouts at a scale sufficient to train complex control policies directly from interaction, enabling years of simulated experience within practical development timelines. This capability has shifted control development away from analytical modeling toward empirical optimization, most notably model-free reinforcement learning~\cite{heess2017emergence}, in settings where contact dynamics, balance, and disturbance rejection are difficult to model explicitly. These approaches have been applied successfully to whole-body locomotion~\cite{hwangbo2019learning,rudin2022learningwalkminutesusing}, contact-rich manipulation~\cite{andrychowicz2018learning}, compliant interaction~\cite{portela2024learningforcecontrollegged, margolis2025softmimiclearningcompliantwholebody}, and agile behaviors such as stepping, crouching, and recovery from external perturbations~\cite{jenelten2024dtc,zhang2026ame2agilegeneralizedlegged,ze2025twist2scalableportableholistic}.

\begin{figure*}[!h]
    \centering
        \includegraphics[width=\linewidth]{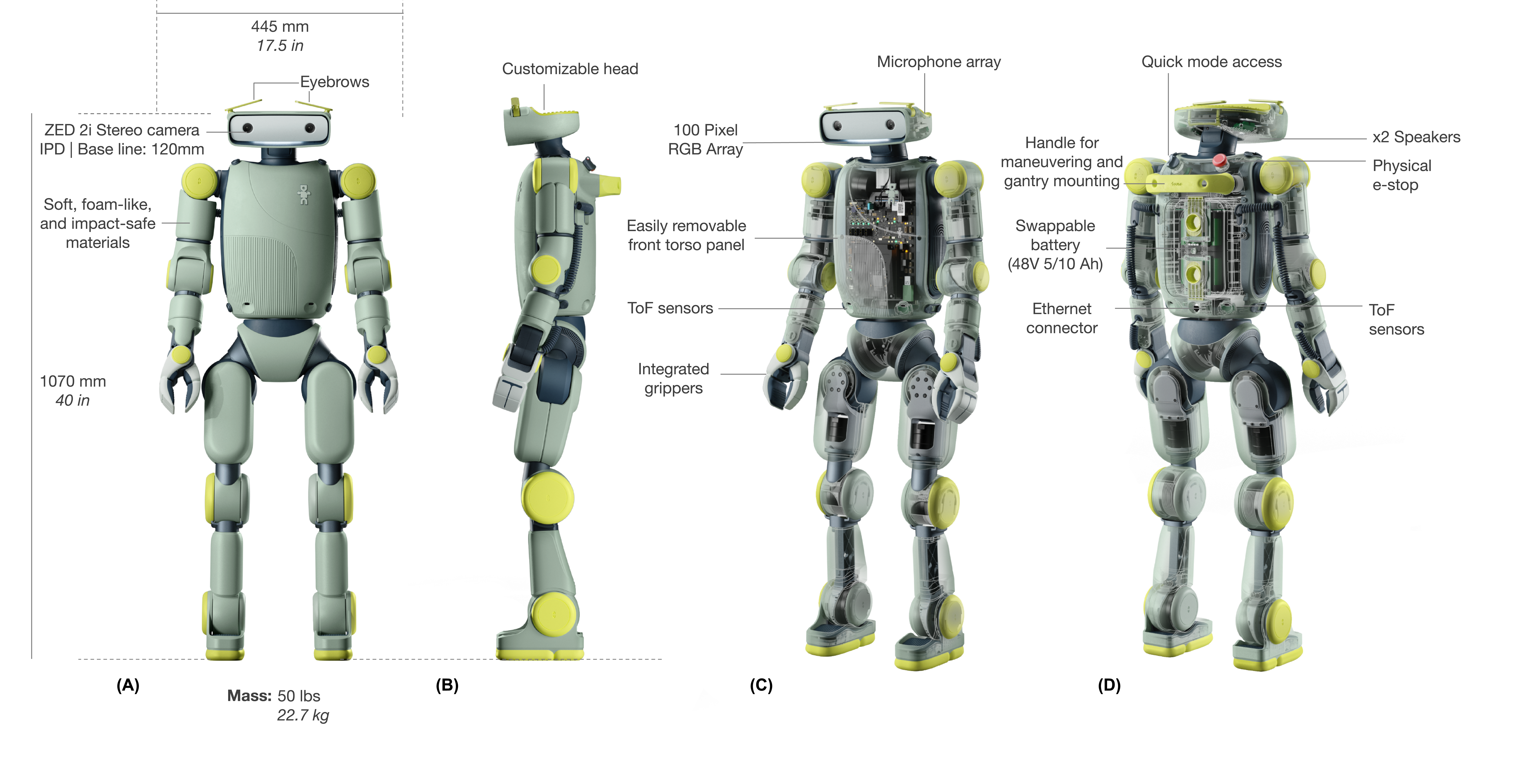}
    \caption{\textbf{Hardware overview.} Key features of the Sprout robot platform from different perspectives: (A) and (B) are true-color renders, (C) and (D) are semi-transparent renders.}
    \label{fig:overview}
\end{figure*}

At higher levels of motor abstraction, learned vision- and language-conditioned systems increasingly generate or modulate motor behavior. One class of these systems is monolithic, in the sense that a single policy couples perception, reasoning, and execution and acts directly on robot actuators. Representative examples include early language-conditioned robotic transformers~\cite{brohan2022rt} and subsequent foundation-model efforts such as Gemini for Robotics~\cite{team2025gemini}, Physical Intelligence~\cite{intelligence2025pi05visionlanguageactionmodelopenworld}, and Generalist AI~\cite{generalist2025gen0}, which primarily demonstrate language-conditioned manipulation often described within the vision-language-action (VLA) paradigm. In these systems, high-level conditioning signals are mapped directly to joint- or end-effector-level commands.

In contrast, many language-conditioned systems developed for humanoid platforms adopt a modular or hierarchical interface, particularly in settings where reliability, interpretability, and verifiability are essential. In applications to manipulation, high-level vision- and language-conditioned models emit commands to explicit low-level control policies rather than acting directly on actuators. Examples include NVIDIA’s N1/N1.5~\cite{nvidia2025gr00tn1openfoundation} and Figure’s Helix~\cite{Figure2025Helix}, which integrate language- and vision-conditioned reasoning with underlying locomotion and manipulation controllers. Related work has also demonstrated language-conditioned whole-body control derived from motion capture or teleoperation data by commanding a low-level control policy~\cite{BostonDynamics2025LargeBehaviorModels}. Other examples of modular control systems decompose autonomy into interconnected but independently validated services---such as perception, mapping, navigation, locomotion, and manipulation---each implemented by specialized algorithms or controllers. For example, recent work uses language models to replace hand-engineered decision logic with tool-calling and program-synthesis paradigms~\cite{ahn2022icanisay, liang2023codepolicieslanguagemodel}, an approach championed by startups including MenteeBot~\cite{Shenkar2025MenteebotApproach} and Flexion~\cite{Flexion2025ReflectV0}.

Across both low- and high-level motor control, this new generation of learned approaches places concrete demands on robotic platforms. At the low level, the success of learned control depends on stable sensing, well-modeled actuator dynamics~\cite{bjelonic2025bridginggapsystematicsimtoreal}, physically accurate robot models, and simulation-deployment parity~\cite{jakobi1995noise,Peng_2018}. At higher levels, supporting the full space of current and emerging approaches requires a platform that enables modular decomposition without obstructing end-to-end methods. Such platforms must also provide essential system services---such as logging, localization, and sensor integration---that are required by both paradigms. We therefore focus on core robot services that can be leveraged or replaced depending on the needs of the developer.

\section{Hardware Design}
Our hardware design is guided primarily by an emphasis on safety, scalability, and aesthetics. We believe that progress toward general-purpose robots operating in unstructured human environments depends on systems that are safe, expressive, and approachable. Sprout’s hardware architecture reflects these priorities, balancing performance, safety, and manufacturability within strict size and mass constraints.

Our approach to safety is structured across three complementary layers. 
The first layer addresses safety at the mechanical and electrical level.
Sprout is intentionally compact and lightweight (see full technical specifications, Supplementary Table~\ref{table:tech_specs}), reducing kinetic energy and enabling safe operation in close proximity to people. The mechanical design minimizes pinch points, hot surfaces, and electrical hazards. Soft, deformable exterior materials further reduce impact forces and support direct physical interaction with people and everyday objects.
The second layer operates at the embedded software level. A dedicated safety subsystem runs on embedded processors independent of the application compute stack. This layer supports real-time monitoring and safety-critical functions, including integration with time-of-flight obstacle sensors and enforcement of system-level constraints even under application-level faults (see Supplementary Figure~\ref{fig:fov}).
The third layer consists of application- and policy-level safety mechanisms. These include compliant motor control policies that limit interaction forces, as well as vision-based systems that support safe navigation and decision-making in human environments. Together, these layers provide defense-in-depth, enabling day-to-day operation with less risk.

Sprout's mechanical design emphasizes low mass and compliance while maintaining the strength and durability required for real-world use. This balance presents significant engineering challenges, which we address by relying heavily on simulation to model and optimize weight, power consumption, and thermal performance across the system. This process enables aggressive mass reduction without compromising structural integrity or reliability. Actuators are selected and tuned on a per-joint basis, favoring appropriately sized motors and gear reductions over oversized components. This approach reduces mass, power draw, and thermal load while improving controllability and safety.

In particular, designing lightweight arms capable of meaningful manipulation at small scales required careful integration of actuation, structure, and wiring. Routing power and signal cables through densely packed articulated joints while maintaining reliability and serviceability proved challenging. To address this, we selected motors early in the design process, and designed the system from the outset to support custom single-degree-of-freedom grippers. These grippers prioritize durability and consistent grasping performance while balancing grip strength against injury risk during interaction; as such, grip force at the fingertips is limited by software to \SI{12}{N}. For arm-payload details, see Supplementary Table~\ref{table:payloads}.

At the core of Sprout’s compute architecture is an NVIDIA Jetson AGX Orin, which provides primary system compute for perception, planning, and high-level decision-making. The Jetson interfaces with a custom electronics motherboard responsible for safe power distribution, real-time sensor acquisition, and deterministic motor control---it also simplifies wiring and assembly (see Supplementary Figure~\ref{fig:compute}). High-level motion commands are transmitted from the Jetson to embedded motor controllers over Ethernet; time-critical control loops then execute locally on those controllers. These controllers communicate directly with actuators using real-time motor control protocols, ensuring bounded latency and deterministic update rates independent of application-level compute load. The same architecture supports deterministic communication with onboard sensors and continuous low-level monitoring of the power system, enabling rapid fault detection and safe responses.

Sprout uses a custom-designed battery optimized for high energy density within tight mass and volume constraints. The battery is built from industry-standard cells and managed by a commercial battery management system, ensuring safety, reliability, and serviceability (details in Table~\ref{table:tech_specs}).

Applying design principles from consumer electronics, we engineered Sprout for manufacturability from the outset.
Assemblies are designed with low part count, standardized fasteners, and clear assembly order, while integrated subsystems are favored over bespoke components. Mechanical and electronic designs are co-developed to minimize wiring complexity, support repeatable cable routing, and enable automated or semi-automated assembly. Key challenges addressed through this approach include maintaining tight tolerances in lightweight structures, managing thermal dissipation in compact enclosures, and ensuring serviceability without increasing assembly complexity. By aligning the design with established consumer electronics manufacturing processes and a flexible global supply chain, Sprout can scale from low-volume builds to mass production while maintaining quality, reliability, and cost control.

Sprout’s aesthetic design draws inspiration from a century of robots in popular culture, emphasizing approachability and expressivity. Facial elements such as an LED array and motorized eyebrows support nonverbal communication, while an integrated audio system enables natural spoken interaction. A four-microphone array supports speech recognition and sound-source localization, and onboard speakers enable expressive audio output. Sprout’s software interfaces allow rapid experimentation with human-robot interaction behaviors, and the head design supports cosmetic customization using standardized interlocking toy brick components.

\begin{figure*}[!h]
    \centering
        \includegraphics[width=\linewidth]{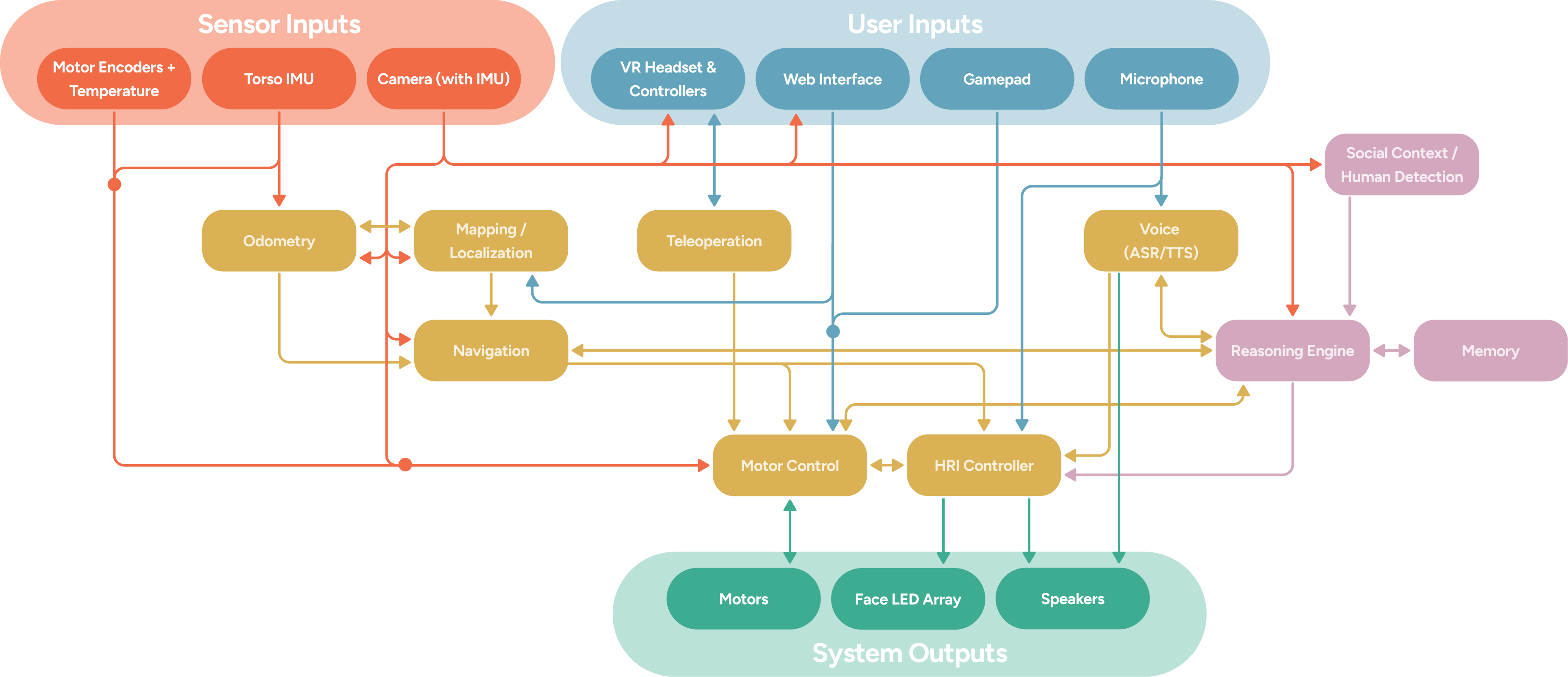}
    \caption{\textbf{Software overview.} A conceptual sketch of the dominant processes and primary channels of communication between them.
    Since the communication follows a pub-sub model, additional edges may exist in practice.
    For example, the web interface could communicate with any service---depicted connections correspond to core capabilities.
    Pink-colored nodes (e.g., reasoning engine, memory, and social context) are included for illustrative purposes, but are not initially available as part of our SDK.
    }
    \label{fig:software_overview}
\end{figure*}

\section{Software Architecture}

\subsection{Overview}

Sprout is built as a component-based AI platform, emphasizing modularity over end-to-end solutions. This design allows us to integrate state-of-the-art technologies across various domains, including perception, planning, locomotion, and manipulation, while ensuring flexibility and ease-of-use. Our aim is to provide developers with a robust foundation that supports rapid iteration as new advancements emerge.

Because components are isolated behind explicit APIs, individual services can be upgraded or replaced without requiring a redesign of the full system. Sprout ships with a complete baseline set of robotics services that enable immediate usability out of the box. Researchers and advanced developers can swap in their own implementations (e.g., a custom perception stack or learned navigation policy) while continuing to rely on the rest of the platform for integration, safety, and deployment.

\subsection{Deployment}

Each service is run in a Docker container, allowing developers to control what is deployed on the robot. Containers can be launched individually or as part of a constellation of services, depending on the application requirements. We provide a GUI to facilitate easy deployment and management of these services, including logical groups of services that should run together (e.g., voice recognition and speech synthesis).

Core platform services related to safety and real-time control run with explicit resource isolation to ensure consistent latency and throughput. In practice, this includes CPU affinity and scheduling policy controls, prioritized execution, and container-level resource limits so developers can deploy additional services without destabilizing the control stack. This approach balances predictability for safety-critical workloads with flexibility for application compute.

\subsection{Communication}

Inter-process communication on the AGX host is primarily handled via ROS 2~\cite{macenski2022robot} for its tooling, ecosystem, and standardized message passing. For high-bandwidth data paths (e.g., images, point clouds), selected pipelines use shared-memory or zero-copy transports to reduce serialization overhead and end-to-end latency. We use RMW Zenoh~\cite{rmwzenoh} to improve performance and to support seamless bridging between on- and off-board compute resources during development (e.g., running a heavy model on a workstation while maintaining the same interface and message contracts).

Additional protocols are used where they best match the constraints of a link or device. For Motor Control Modules (MCMs), we use CBOR~\cite{cbor} over Ethernet, optimized for compact, deterministic control messaging. The teleoperation headset uses JSON over WebSocket for control and state exchange, with WebRTC for low-latency video and audio streaming. For visualization tools, we use the Foxglove WebSocket protocol for structured telemetry and playback.

\subsection{Developer Experience}

We provide a companion web application and integration with the Foxglove visualization platform~\cite{foxglove} to make Sprout observable and debuggable during development and field testing. The platform includes a structured logging pipeline that captures synchronized telemetry, events, and sensor streams for offline inspection and replay. These tools surface key runtime signals (system state, perception inputs, navigation status, localization), enabling fast iteration without requiring deep knowledge of the underlying service topology.

The SDK exposes stable APIs for building on top of Sprout. Today, this consists of ROS 2 message definitions and example code for interacting with services in Python and C++. Where appropriate, we include reference implementations for common patterns (request/response, streaming telemetry, action-style interfaces) and provide templates for packaging custom services as containers so they can be deployed and managed alongside the baseline constellation. Future releases will include additional language bindings and non-ROS APIs to broaden accessibility.

At launch, we provide end-to-end examples for common workflows, including:
\begin{itemize}
    \item Deploying and running a custom low-level locomotion policy
    \item Using voice commands to navigate the robot via LLM-based agents
    \item Recording teleoperation sessions for analysis and playback
\end{itemize}

\begin{figure*}[!h]
    \centering
        \includegraphics[width=\linewidth]{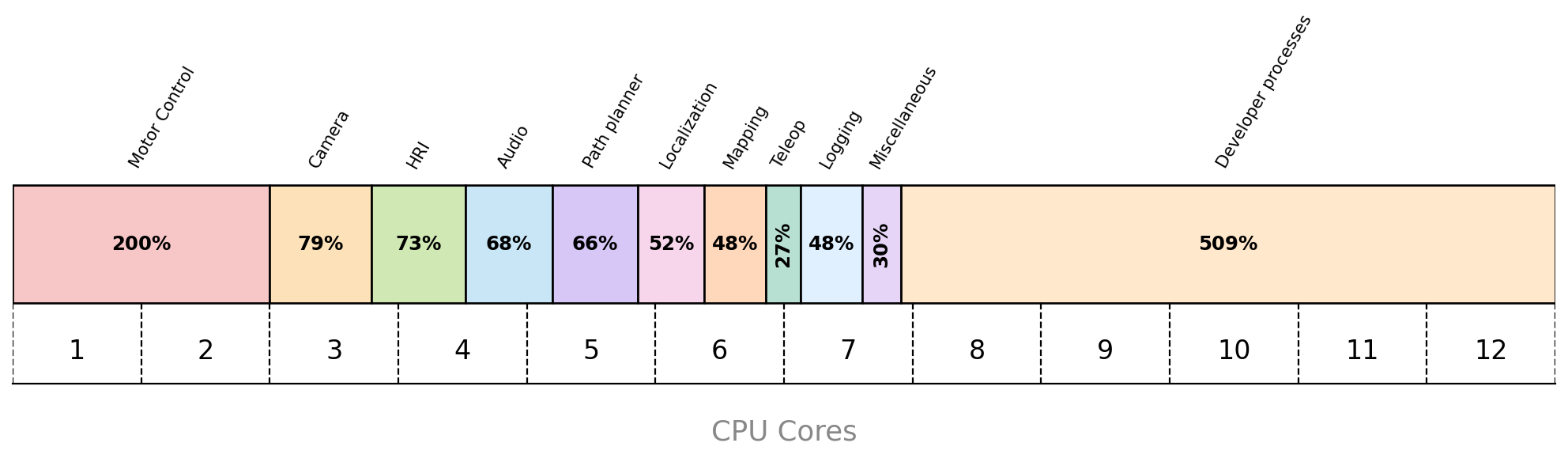}
    \caption{\textbf{CPU utilization.} Approximate snapshot of the CPU demands of core onboard software services. In addition, some services leverage the GPU (GPU profiling not shown).}
    \label{fig:cpu_cores}
\end{figure*}

\subsection{Performance Optimization}

To maximize available compute for user applications, we optimize the baseline stack for predictable latency and efficient resource use. Key strategies include minimizing internal message-passing overhead in hot paths, moving performance-critical components to C++ (with Python bindings where ergonomics matter), and using shared-memory and zero-copy transports for large data.

\section{Core Software Services}

\subsection{Motor Control}

The motor control system is designed to balance developer ergonomics with the requirements of safe, reliable operation in human environments. Concretely, this means the system must be composable and debuggable, while remaining robust to modeling error, contact variability, and external perturbations. To this end, we avoid relying on a single monolithic controller. Instead, we favor structured control interfaces that expose clear abstractions---commands, modes, transitions, and safety envelopes---so that learned components can be deployed with explicit constraints and guarantees.

\subsubsection{Orchestration via State Machines}

\begin{figure}[t!]
  \centering
  \includegraphics[width=\linewidth]{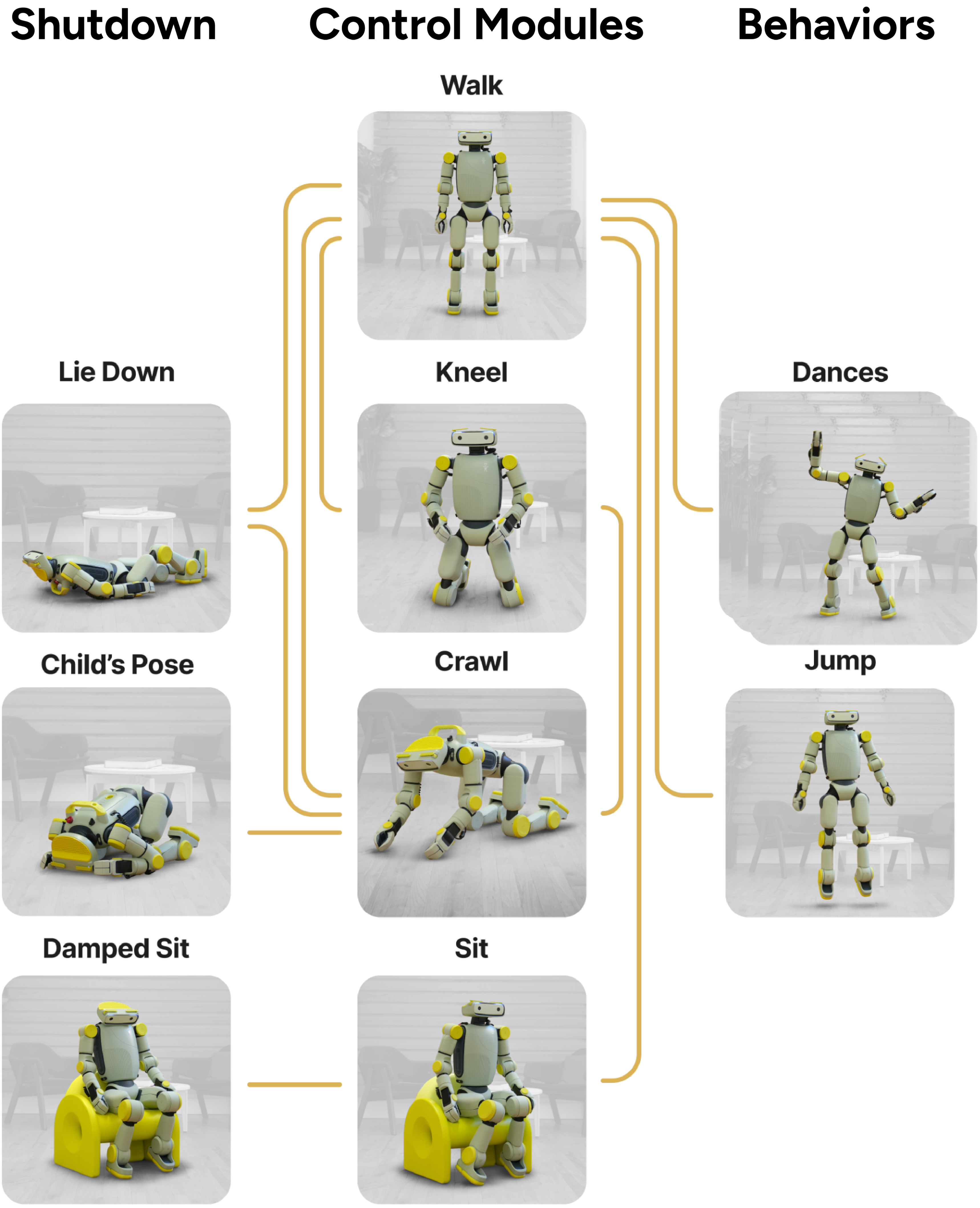}
  \caption{\textbf{Locomotion system.} Supported transition structure between motor modules and motor behaviors.}
  \label{fig:motor_control}
\end{figure}

To meet these goals, control is organized around a finite state machine that orchestrates a set of discrete control modes (Figure~\ref{fig:motor_control}). Each mode encapsulates a well-defined behavior (e.g., standing, walking, kneeling), along with mode-specific safety checks, validity conditions, and transition logic. While end-to-end policies can achieve strong performance in unconstrained settings, the platform adopts a state-machine-based architecture that allows for explicit definition of valid operating regimes, mode-local safety mechanisms (e.g., posture bounds, joint limit monitoring), controlled transitions with preconditions and recovery behavior, and isolation of changes, so modifications to one mode do not destabilize others. The state machine is exposed through a programmatic API, enabling external systems (teleoperation, autonomy stacks, testing tools) to request mode transitions while inheriting the built-in safety logic.

\subsubsection{Control Modes}

Each control mode is backed by one or more reinforcement learning policies trained to satisfy a parameterized command interface. Commands specify desired physical objectives---such as base linear and angular velocity, root orientation, height from the ground, or joint-space targets---rather than raw actuator signals.

Policies map short histories of proprioceptive observations, inertial measurements, and previous actions to intermediate control targets. These outputs are interpreted and constrained by lower-level mechanisms, including PD control, current limiting, and power constraints. This separation ensures that learned behavior remains bounded by hardware-safe execution layers.

All policies are trained in IsaacSim using training infrastructure built on IsaacLab~\cite{mittal2025isaac}. Each control mode is associated with one or more dedicated policies, enabling targeted training, evaluation, and iteration. This modular structure allows individual behaviors to be replaced or improved independently, without requiring system-wide retraining.

Control modes correspond to qualitatively different behaviors, including walking, kneeling, and crawling. Each mode reflects a distinct set of assumptions about contact configuration and allowable motion, and is trained and validated independently. The walking control mode, designed to support whole-body teleoperation, provides a representative example of how learned control is structured and deployed. In this mode, the operator commands a continuous set of task-level variables describing the desired motion of the robot. These commands include the desired base linear velocity, the desired yaw rate, the desired root orientation specified by roll and pitch, the desired root height, and target configurations for upper-body joints such as the arms and torso.

Learned policies map these commands, together with proprioceptive state, to control targets that realize the requested motion. This design allows the operator to control high-level motion intent directly, without micromanaging details related to movement dynamics. These commands also execute seamlessly atop the compliant motion policies underneath, ensuring safety even under human control. Importantly, the same command interface can be reused across simulation, teleoperation, and higher-level autonomy stacks.

\subsubsection{Transitions}

Transitions between behaviors are a common source of instability in robotic systems, particularly when control modes are trained independently and rely on different assumptions about contact state, posture, or admissible motion. Abrupt switching between policies can induce discontinuities in commanded motion, violate safety constraints, or excite unmodeled dynamics, even when the individual control modes are stable in isolation.

We therefore treat transitions between control modes as unique components of the control system. A transition may consist of one or more policies executed sequentially to safely move the robot from the state distribution of one mode to that of another. Rather than assuming that a single policy can reliably span both regimes, transitions are explicitly structured to manage changes in contact configuration, body posture, and control objectives.

Transition behaviors are implemented as learned tracking controllers trained via imitation learning. Each transition consists of a short motion sequence that brings the robot into a configuration compatible with the destination control mode while respecting safety constraints throughout. Training data for these sequences is drawn from a combination of human motion capture and animated trajectories. To ensure generalization and robustness to small variations across deployments, policies are initialized randomly by sampling from a broad distribution around the nominal trajectory. 

As with control modes, transitions define their own validity and safety conditions, which may differ from those of the source or destination modes. These transition-aware constraints allow the system to monitor posture, contact state, and actuator limits throughout the transition process and to prevent or abort transitions that would lead to unsafe configurations. This approach enables smooth and reliable behavior across mode boundaries.

\begin{figure*}[t] 
  \centering
  \includegraphics[width=\linewidth]{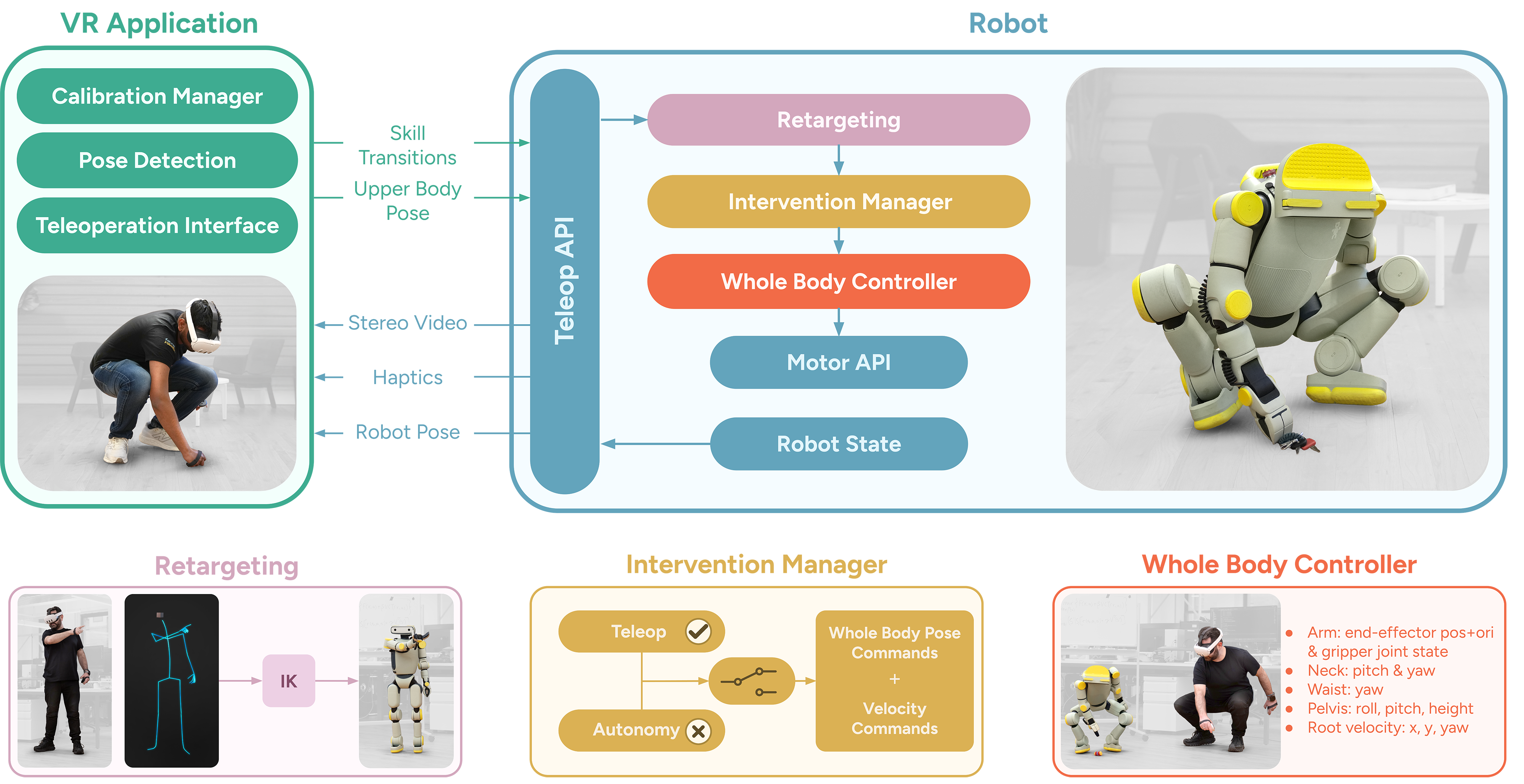}
  \caption{\textbf{Teleoperation system.} The communication between the VR system (green) and robot services as mediated by the core robot APIs (blue), with schematic breakouts depicting the retargeting step (pink), switching between teleoperation and autonomy (yellow), and the whole-body control layer (orange).}
  \label{fig:teleop}
\end{figure*}

\subsubsection{Compliance}

Physical interaction is unavoidable in human environments, and compliance is a necessary property for safe and robust operation. In this context, compliance refers to the ability of the robot to yield appropriately to external forces, dissipate energy during contact, and avoid generating large or impulsive interaction forces in response to disturbances.

The platform treats compliance as a fundamental control property rather than a mode-specific feature. All control modes, and certain transitions, are trained and executed under constraints that promote compliant behavior, encouraging controlled, non-aggressive responses to a wide range of incidental contacts. In practice, this means that unexpected interactions such as human contact, environmental collisions, or external perturbations do not elicit unstable or forceful corrective actions.

Compliance is achieved by training policies in environments that expose the robot to disturbances and modeling uncertainty. This is paired with a reward structure that encourages behaviors that remain stable under perturbation and minimally counteract externally-applied forces.

Beyond safety, compliance expands the space of feasible human-robot interactions by enabling closer proximity, physical guidance, and shared environments. Importantly, this is achieved without fragmenting the control architecture or introducing special-case logic, preserving the modularity and predictability of the overall system.

\subsubsection{Sim-to-Real Calibration}

Reliable whole-body control and learning depend on a shared understanding of the robot's physical state and capabilities. In practice, many failures in robotic systems arise not from deficiencies in control algorithms, but from mismatches between assumed and actual hardware behavior~\cite{jakobi1995noise,bjelonic2025bridginggapsystematicsimtoreal}. Sim-to-real gaps such as calibration errors or unmodeled actuator dynamics accumulate silently, undermining reproducibility and limiting the usefulness of learned controllers.

We employ several industry-standard approaches to ensure precise calibration. At the level of individual motors, we perform zero-referencing using a precise calibration table to enforce a known physical configuration and ensure consistent encoder-to-geometry mapping. Similarly, we perform extrinsic calibration of the camera using fiducial markers and a self-calibration routine to ensure accurate visuo-spatial perception. Calibration primitives are exposed through the platform API, allowing higher-level systems to leverage information about robot state and configuration. To support reproducibility, we release zeroing specifications that enable in-house and third-party calibration workflows to follow the same conventions.

\subsubsection{Actuator Models}

Beyond encoding accuracy, actuator dynamics are often the dominant factor governing control performance and sim-to-real transfer~\cite{hwangbo2019learning,bjelonic2025bridginggapsystematicsimtoreal}. To address this, we provide actuator models using DC motor dynamics augmented with delay, saturation, and power constraints, capturing several key non-idealities present in real systems. We refine our actuator models using motor manufacturer specifications, dynamometer measurements, and optimization from real data. Similarly, system-level current and power limits are explicitly represented to ensure motors operate within feasible and safe bounds given electrical constraints. All actuator assumptions are included in the platform documentation to make these constraints transparent to developers.

\subsubsection{Supported Description Formats}

Finally, to support consistent physical modeling across tools and workflows, we provide several standard robot description formats, including USD, URDF, and MJCF. IsaacSim~\cite{mittal2025isaac} serves as the primary simulation backend, though our robots remain compatible with MuJoCo-style tooling~\cite{todorov2012mujoco}. We make these formats available as a part of our SDK.

\subsection{Whole-Body Teleoperation}
\label{sec:teleoperation}

Using our compliant whole-body control system as a foundation, we developed a VR-based teleoperation interface for full-body control.
Recent work has demonstrated the effectiveness of VR-based teleoperation for humanoid whole-body control~\cite{li2025amo},
enabling operators to provide intuitive commands for dexterous loco-manipulation tasks.
This section describes each of the components of our teleoperation system (Figure~\ref{fig:teleop}).

\subsubsection{Embody: User-Friendly Application for Meta Quest Devices}

We developed Embody, a user-friendly Unity application for Meta Quest devices that enables full-body
teleoperation of Sprout. Similar to other VR-based teleoperation systems for manipulation~\cite{iyer2024openteach},
our application leverages the affordability and accessibility of consumer VR headsets to provide an intuitive
control interface. Embody provides users with an interactive experience for controlling
the robot through a series of intuitive menus, heads-up displays, and control mappings on the Quest hand controllers.
The application uses Meta's Movement SDK and sends body pose keypoints to the robot, while a backend service running on the robot sends visual and haptic feedback to the user.
Embody runs a calibration routine at startup to estimate
the user's arm lengths, torso height, and comfortable range of vertical motion (for mapping squatting) for isomorphic retargeting to the robot.
This maps the user's morphology in a custom way to the robot's size and shape, and provides a natural feeling to control of its motions.
The control defaults to full-body control, where the user can lean forward and squat to command the robot close to the ground to reach even small objects like keys or paper with ease. An upper-body-only and seated mode are available as well.
Embody also provides options to switch the robot to modes such as kneeling and sitting while preserving the ability to teleoperate the upper body in these situations (as well as waist pitch for bending over when kneeling).
All of this combines to provide the user a commanding range of motions for reaching objects on or near the ground.
The application manages different states of operation---Calibration, Teleoperation, Mode
Selection, Data Collection, and Autonomy, through a state machine that also can change control mappings or other user preferences based on the different mode of operation.

\subsubsection{Whole-Body Policy}

To produce whole-body behavior, we extend our compliant whole-body policy to support teleoperation.
The policy controls pelvis pitch, roll, and height while maintaining compliance on the upper body.
To support matching commanded pitch and roll, we use projected gravity (the gravity vector expressed in the robot's body frame) as the control signal. By commanding the policy to track this projection rather than absolute orientation, the robot maintains stable body poses relative to gravity regardless of terrain variations. For height control, we train the policy to match a height command corresponding to the desired torso-to-ground distance.

\subsubsection{Retargeting}

We transform wrist keypoints from the pelvis-relative frame to the robot's frame of reference, scaling
the received positions using factors computed during calibration. For full-body retargeting, we linearly
map the calibrated standing and crouching heights to the policy's minimum and maximum height ranges, with
pitch and roll similarly mapped from VR input ranges to the policy's allowable values. The retargeted
Cartesian poses are then converted to joint angles using the PINK inverse kinematics library~\cite{pink2024},
with the solver tuned using position cost, orientation cost, and regularization cost to avoid singularities.

\subsubsection{Data Collection System}

We developed a data collection system alongside the teleoperation system that allows users to annotate,
record, and play back demonstrations. The system logs stereo RGB images at \SI{30}{Hz}, proprioceptive data
including whole-body pose commands (end-effector Cartesian poses, gripper positions, pelvis pitch, roll,
and height commands), joint position commands, and velocity commands at \SI{50}{Hz}, as well as proprioceptive
states at \SI{125}{Hz}. The data collection system also allows users to annotate demonstrations with a simple button press, in order to efficiently segment out usable trajectories for training from other motions. Recordings and annotations can also be managed from the app.

Beyond pure demonstration collection, the system supports DAgger-style interventions~\cite{ross2011dagger} following recent work in manipulation learning~\cite{intelligence2025pistar06}
to address the covariate shift problem inherent in behavioral cloning. During autonomous policy execution,
an operator monitors the robot's behavior through the VR interface and can seamlessly take over control
when the policy encounters states outside its training distribution or begins to fail. These intervention
segments are logged alongside the autonomous rollouts, capturing expert corrections precisely at the
failure modes that matter most. Our system allows the user to pause policy playback at the point of an error, at which point we project the current pose of the robot at that point as a ``ghost'' controller to which the user can align their controllers when they assume control. When the user wants to demonstrate a motion intervention starting at the paused position, the user can then align their controller to these ghosts and start collecting new data from the exact position and orientation where the robot left off. 
This provides clean, novel examples that can be aggregated as on-policy data with the original demonstrations for retraining. 
This iterative process helps policies become more robust to compounding errors, which is critical for learning reliable whole-body manipulation behaviors.

\subsection{Mapping \& Navigation}
\label{sec:map_nav}

\begin{figure*}[t!]
  \centering
  \includegraphics[width=\linewidth]{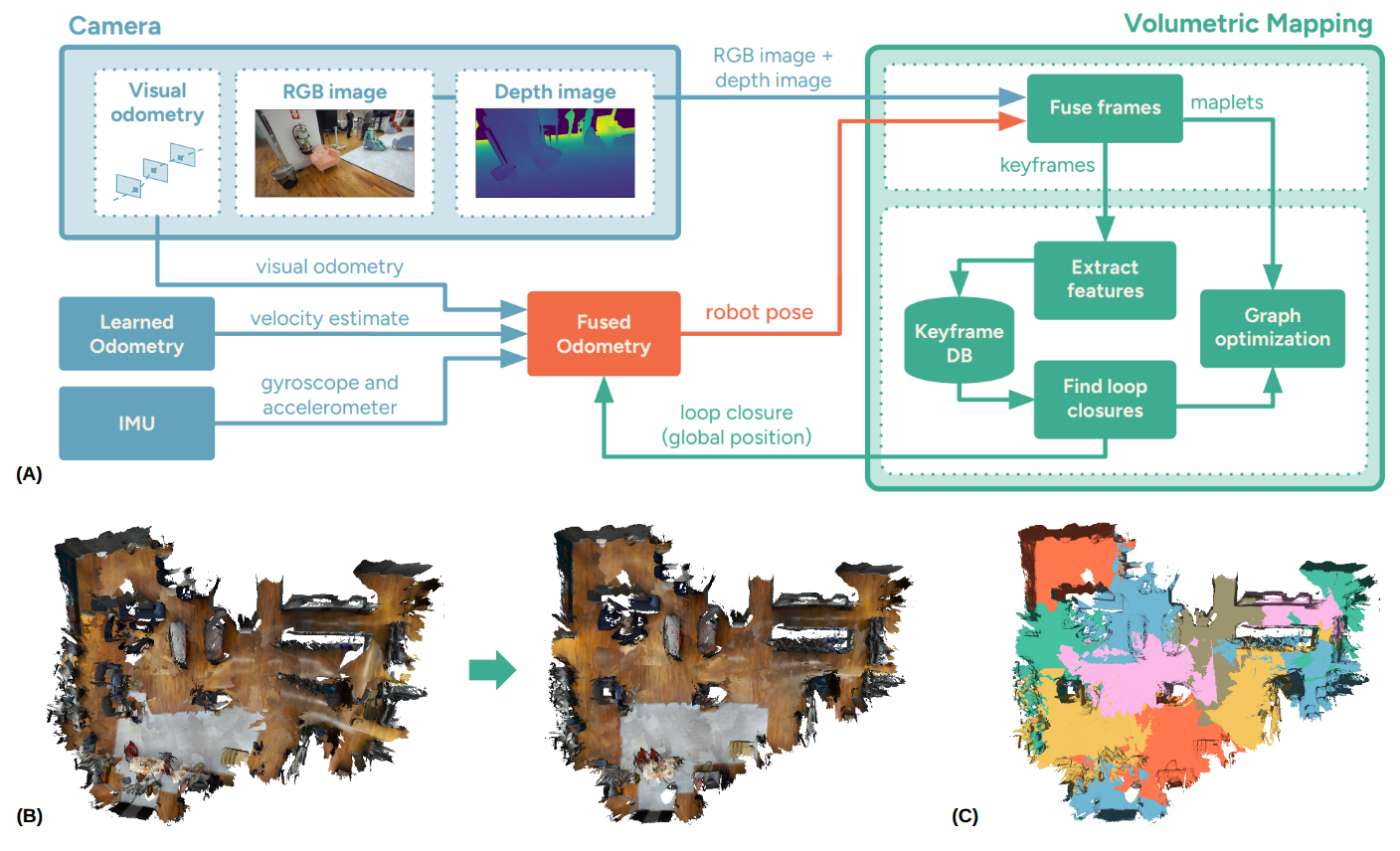}
  \caption{\textbf{Mapping and localization.} (A) An overview of the sensor inputs (blue), the \emph{Fused Odometry} module (orange), and the \emph{Volumetric Mapping} algorithm (green); these components work together to estimate the robot's state and environment. (B) A map before and after graph optimization. (C) The same map showing its individual maplets in different colors.}
  \label{fig:mapping}
\end{figure*}

Sprout's indoor autonomy system consists of three core components: a custom fused visual-inertial-kinematic odometry pipeline designed for bipedal locomotion, a lightweight on-demand mapping system with a low compute footprint accessible via both CLI and GUI, and a modular navigation stack that integrates pose tracking, local obstacle avoidance, and global route generation. The following sections describe each component in detail.

\subsubsection{Odometry}
We developed a \emph{Fused Odometry} system that combines visual, inertial, and kinematic information to produce a low-latency, high-frequency state estimate suitable for agile bipedal locomotion. The estimator is implemented as an Extended Kalman Filter (EKF) that integrates measurements from the ZED2i stereo camera, the onboard inertial measurement unit (IMU), a learned proprioceptive velocity estimator from the 
robot's motor control policy, and, when available, global position information from loop closures (Figure~\ref{fig:mapping}{}A). The filter outputs pose, velocity, and associated covariance estimates at \SI{50}{Hz}.

Kalman filtering remains a standard approach for state estimation in legged robots due to its computational efficiency and ability to incorporate heterogeneous sensor modalities. Prior work demonstrates the benefits of tightly fusing inertial, visual, and leg kinematics to reduce drift and improve robustness in challenging scenarios~\cite{camurri2020pronto, yang2023cerberus}. Our system builds directly on this line of work while adapting the fusion strategy to the specific characteristics of Sprout’s morphology, gait dynamics, and sensing configuration.

In contrast to wheeled---and even quadrupedal---platforms, Sprout exhibits intermittent, asymmetric foot contacts and rapid changes in support surfaces during walking. These characteristics introduce additional noise in kinematic measurements and complicate contact inference. 
Our approach mitigates these challenges by leveraging redundant visual and inertial cues while using the learned estimator to regularize short-term motion estimates. The resulting odometry is robust to texture-poor indoor environments, rapid body motions, and depth sensing noise, forming a stable foundation for downstream mapping and navigation.

\subsubsection{Mapping (and Localization)}
Indoor SLAM (Simultaneous Localization and Mapping) for legged robots presents distinct challenges due to drift-prone odometry, rapid six-degree-of-freedom body motions, intermittent and uncertain foot contacts, and possible degradation of depth or visual information in texture-poor environments. Classical visual SLAM systems such as ORB-SLAM3~\cite{campos2021orbslam} and related monocular or RGB-D pipelines excel on wheeled robots, where motion is smoother and exteroceptive observations are more stable. Likewise, 2D LiDAR-based indoor SLAM approaches~\cite{zhang20242dliwslam} and mobile-robot mapping systems assume near-planar motion and consistent sensor geometry. In contrast, recent work on legged-robot state estimation and mapping emphasizes the importance of tightly coupling inertial, kinematic, and visual cues to maintain robustness under aggressive motions and variable ground interaction~\cite{camurri2020pronto, yang2023cerberus, xiao2025geoflowslam}. These efforts underscore the need for SLAM architectures that tolerate transient perception failures while maintaining global map consistency in 3D environments.

Our SLAM pipeline, \emph{Volumetric Mapping}, builds on these insights by constructing a dense 3D representation of the environment using a Truncated Signed Distance Field (TSDF), a volumetric grid where each voxel stores the signed distance to the nearest surface, truncated to a finite range for computational efficiency. TSDF fusion naturally denoises depth measurements, provides smooth and 
continuous surface estimates, and supports both local planning and global reconstruction. Depth images from Sprout's RGB-D camera are fused into TSDF volumes at a high rate, forming the basis for our dense mapping layer.

A key architectural feature of the system is its division of the world into locally consistent volumetric submaps, or \emph{maplets} (shown in Figure~\ref{fig:mapping}C). Each maplet represents a rigid, non-deforming coordinate frame in which TSDF fusion occurs. New maplets are created as the robot moves beyond the predefined spatial extent of the current maplet (to mitigate the effects of long-term drift), or when degradation in odometry quality is detected (e.g., a sudden jump in estimated robot pose). This design prevents reconstruction errors---arising from inaccurate depth measurements, rolling-shutter distortion, or degraded odometry---from corrupting the global map, while keeping the local fusion strategy computationally efficient. It also mirrors the submap strategies used in other successful dense and semi-dense SLAM systems~\cite{yan2024survey}, where locality is exploited to maintain real-time performance while enabling scalable global optimization.

Global alignment is handled through a hierarchical pose graph comprising both keyframe poses and maplet poses. Odometry constraints derived from our fused estimator, along with visual loop closures, introduce spatial relationships between maplets that allow the back-end optimizer (implemented in GTSAM~\cite{dellaert2012factor}) to correct accumulated drift. Loop closure constraints between revisited regions produce non-local updates that 
adjust the arrangement of maplets in the global frame while preserving their internal rigidity. This strategy achieves a balance between computational efficiency and global consistency comparable to multi-layer factor-graph approaches in recent legged-robot SLAM systems~\cite{xiao2025geoflowslam} (Figure~\ref{fig:mapping}{}B).

From a systems perspective, \emph{Volumetric Mapping} is designed to achieve strong reconstruction and loop-closure consistency under tight onboard compute budgets. In our internal benchmarks, it outperforms RTAB-Map~\cite{labbe2019rtabmap} while using approximately 30\% of its compute, enabling onboard operation with low CPU load (typically a third of a single core as shown in Figure~\ref{fig:cpu_cores}).
This efficiency comes from the TSDF fusion pipeline optimized for high-rate integration, together with a maplet-based architecture that keeps fusion locally consistent under drift and enables sparse, asynchronous global optimization.
We further reduce overhead with a custom ML-based loop-closure service that runs asynchronously alongside the real-time mapping pipeline. The loop-closure service uses a cascaded set of neural networks for visual place recognition, feature detection, and feature matching, producing low-latency loop constraints for the pose graph.

This mapping architecture is particularly well suited to Sprout’s locomotion dynamics and sensing configuration. Rapid 6-DoF motions and intermittent foot contacts often lead to short-term degradation in depth or feature tracking; by limiting fusion to compact, rigid maplets and deferring global adjustments to an asynchronous optimizer, the system remains robust to transient noise and maintains a stable mapping backbone even during fast maneuvers. The resulting SLAM pipeline provides reliable, high-fidelity maps tailored to the demands of agile indoor bipedal navigation.

\begin{figure}[t!]
  \centering
  \includegraphics[width=0.445\textwidth]{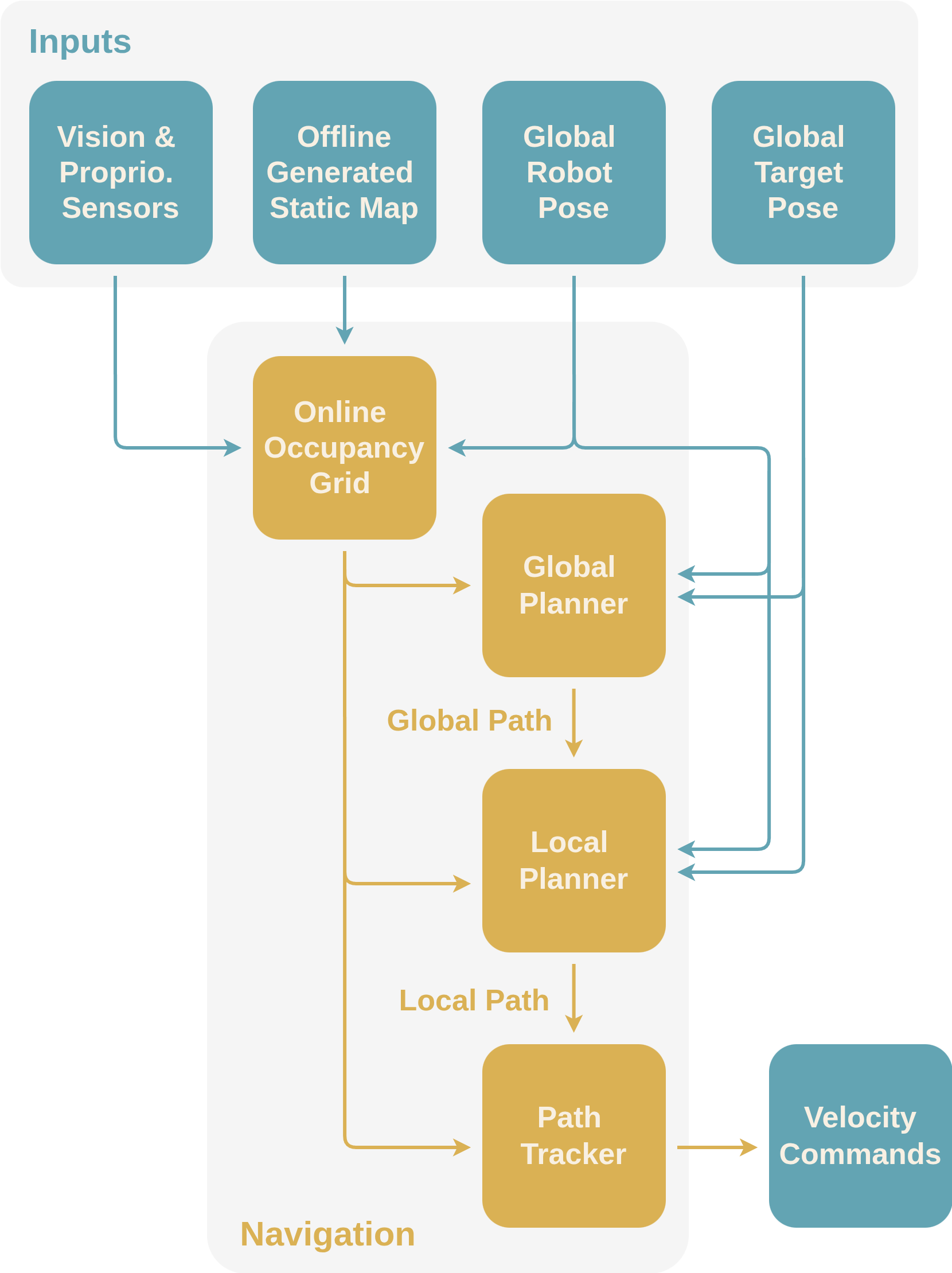}
  \caption{\textbf{Navigation system.} Inputs: an offline-generated static map, global robot pose, global target pose, and vision/proprioceptive sensor data. Outputs: a global path to the target, a local path along the global path, and velocity commands for path following.}
  \label{fig:navigation}
\end{figure}

\subsubsection{Navigation}

We implemented a lightweight navigation system that coordinates locomotion to both static and dynamic target poses. The system is optimized for real-time performance using only onboard compute. The navigation system consists of three main modules: online occupancy grid generator, path planner (global + local planner), and path tracker. Figure~\ref{fig:navigation} shows an overview of the navigation system.

The navigation system operates based on an occupancy grid representation of the environment. The occupancy grid consists of a static layer created by the mapping module and a dynamic layer created online using an OctoMap-based approach~\cite{HornungWBSB13}. The two maps are then fused to create the final occupancy grid used for navigation.

Then, a path planner generates the navigation path. The path planner has a global and local component. For both the global and local planners, we employ a custom Hybrid A$^\star$ planner~\cite{dolgov2008practical} to generate paths. Given the up-to-date map of the environment, the global planner generates a path from the robot's current pose to the target pose. Then, the local planner selects a local target along this global path and generates a path to this local target. The global path is updated only when the robot deviates significantly from the global path or if it becomes invalid due to changes to the global map, while the local path is updated every planning cycle (the default planning cycle is \SI{10}{Hz}). Once the local path is generated, a pure-pursuit-based path tracker~\cite{MacenskiS0G23} is used to generate velocity commands to follow the generated path, which are sent to the locomotion system for execution.

\begin{figure*}[t] 
  \centering
  \includegraphics[width=\linewidth]{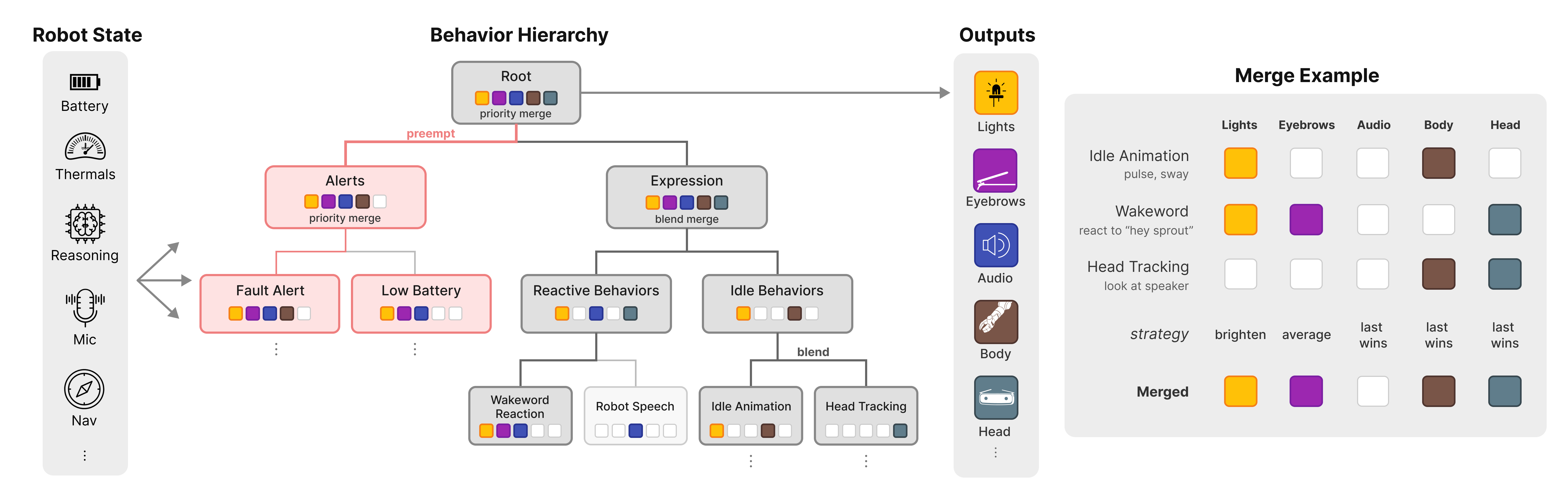}
  \caption{\textbf{Human-Robot Interaction subsystem.} Behavior-generation tree (left) with merge at every level. Leaves emit sparse slot proposals (show elements, audio, motor control); internal nodes blend their children. At the root, alerts preempt other behaviors while expression and idle layers blend. The merge example (right) shows per-slot strategies (brighten/replace/last-wins) producing a single command stream at every level, ultimately sent to hardware.}
  \label{fig:hri_controller}
\end{figure*}

\subsection{Human-Robot Interaction}

The Human-Robot Interaction (HRI) subsystem communicates system status and conveys expressive intent through Sprout's lights, sounds, and mechanical movements (eyebrows, head pose, and upper-body gesticulation). Developers and creators can control these hardware channels alongside the built-in behaviors to design custom interactions without losing system cues. This feedback improves trust and engagement by providing legible indicators for the robot's states and actions.

The HRI service consumes a wide range of robot state spanning system readiness (e.g., power state, battery status), user interaction (e.g., reasoning state, microphone status), and activity of other services (e.g., navigation state). These signals are mapped into coherent indicators that set expectations, reduce confusion, and clearly surface safety-relevant conditions without sacrificing expressiveness in everyday interaction. Examples include warning indicators for high motor temperatures (audible beeping and pulsing orange lights) and a ``thinking'' animation while the reasoner is processing (spinning lights on the face).

Outputs are generated by a hierarchical graph of computation nodes (Figure~\ref{fig:hri_controller}), each of which interprets a subset of state and proposes an action in one or more domains. Rather than emitting a monolithic ``behavior,'' each node produces a partial command structured into independent slots (e.g., LED patterns, audio cues, body and head targets, show-element poses). On each control tick, eligible nodes run, emit their slot outputs, and propagate these proposals upward; intermediate nodes then combine child proposals into a single, coherent command using an explicit merge policy. This slot-based structure allows domains to be coordinated tightly (such as synchronizing an eyebrow motion with a light cue) while remaining decoupled enough that unrelated channels can continue unaffected. Behaviors can be authored from manual keyframe-based animations, teleoperation recordings, learned whole-body controllers, or heuristic generators, all integrated through the same slot-based command interface and tooling pipeline.

Unlike winner-take-all control structures (e.g., state machines), the framework supports multiple merge operators to combine child outputs: additive blending for layered LED effects, select-one arbitration when exclusivity is required, and per-slot last-writer-wins composition when later behaviors should override earlier ones. Blending enables explicit prioritization for high-priority safety behaviors (e.g., critical faults, imminent power loss, collision avoidance) to reliably preempt lower-priority expression, while interactive and idle behaviors can still contribute when safe.

\subsection{Conversation and Reasoning}
\label{sec:reasoning}

Our platform, at present, does not provide a turnkey conversational agent for autonomous operation. Instead, it exposes a suite of core robot services that developers can assemble into their own agent-based systems. These services include ROS 2 topics for event and state signaling, as well as a Model Context Protocol (MCP)~\cite{MCP} server that hosts a variety of tools for agentic control. Together, these communication channels and tools can be orchestrated by LLM-based agents to perform complex, end-to-end reasoning tasks. This modular architecture allows developers to design agents tailored to their specific applications while taking advantage of the growing ecosystem of agent frameworks with first-class MCP support.

\begin{figure*}[t] 
  \centering
  \includegraphics[width=\linewidth]{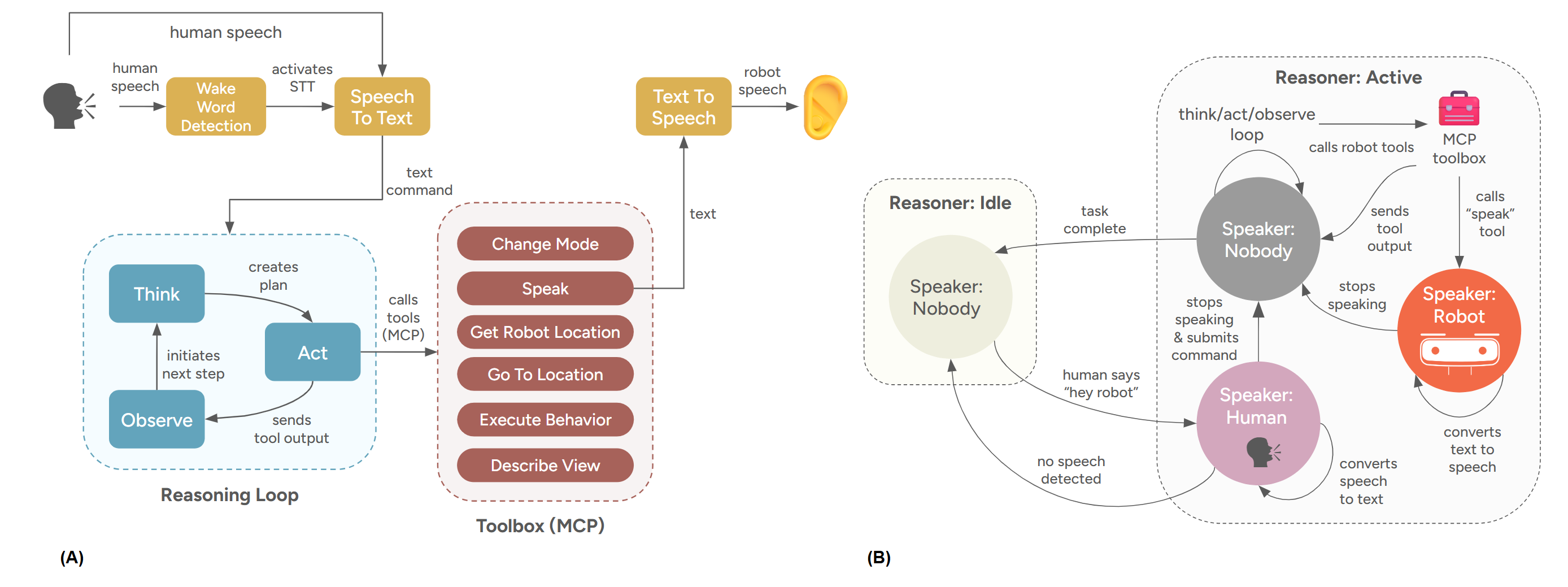}
  \caption{\textbf{Conversation and reasoning.} (A) General flow of agentic, conversational interaction, including the core Think-Act-Observe reasoning loop. (B) State machines for both reasoning and conversational turn-taking, useful for triggering automatic HRI responses, such as listening, speaking, and thinking.}
  \label{fig:reasoner}
\end{figure*}

Conversational closed-loop interactions (Figure~\ref{fig:reasoner}{}A) rely on three core components: a wake-word detector, an automatic speech recognition (ASR) service, and a text-to-speech (TTS) tool. The wake-word detector, built on openWakeWord~\cite{openWakeWord}, continuously listens for the activation phrase \textit{“hey robot”} and initiates an interaction---typically by triggering ASR. Once active, the ASR system transcribes spoken input into text that can be processed by an LLM-based reasoning module. The TTS tool, whether called directly or invoked by an agent, then generates verbal responses that support natural, dynamic interaction. In typical workflows, the robot provides spoken feedback after completing a task, reporting results or describing any issues encountered.

We support multiple ASR and TTS providers but have the tightest integrations with Deepgram Flux~\cite{DeepgramFlux} and ElevenLabs v2~\cite{ElevenLabsV2}, which provide high-quality voice I/O with minimal load on the robot’s onboard computer. For developers who prefer these services to run locally, the system can be configured to use NVIDIA Riva~\cite{NVIDIARiva} for both ASR and TTS. 

Beyond conversation, the MCP server provides tools that allow agents to control the robot's core motor functions (Figure~\ref{fig:reasoner}{}B). Agents can switch between motor control modes---such as walking, crawling, dancing, sitting, and more---using dedicated tool calls. The platform also exposes higher-level behaviors that coordinate motion, lighting, and sound, including actions like head nodding, head shaking, high-fives, and handshakes. Agents can query the list of available behaviors at runtime, as it dynamically depends on the current motor mode. Navigation is similarly controllable through MCP tools: agents can query the robot's current position and set navigation goals for autonomous locomotion within a mapped environment.

Together, these capabilities give agents comprehensive control over the robot's interaction, motion, and navigation. Developers can further expand agentic functionality by creating custom tools and binding them to agents via preferred frameworks. Additionally, as the platform continues to mature, we plan to expand the library of tools and services, further increasing the robot's autonomy and enriching its interactive capabilities.

\section{Discussion}
\label{sec:discussion}

Sprout is a humanoid platform designed for safe, expressive, and sustained operation in close proximity to people, with an emphasis on durability and scaled manufacturability. The platform provides integrated support for whole-body behaviors such as walking, kneeling, crawling, and compliant interaction, along with teleoperation, mapping, navigation, and expressive HRI primitives. These capabilities are exposed through a modular software stack that includes stable control interfaces, containerized services, and tooling for deployment, monitoring, and data collection. As additional core capabilities (e.g., visually guided locomotion, autonomous manipulation, etc.) are developed and tested, they can be provided via software updates.

In optimizing the platform, we made a set of deliberate trade-offs guided by reliability and safety. For the initial release, we opted for simpler one-degree-of-freedom grippers rather than multi-fingered hands, prioritizing robustness and low mass. The intended use of the robot emphasizes object fetching, hand-offs, and physical interaction in shared spaces, as well as the ability to fall, crawl, and recover without damaging delicate end effectors. While increased dexterity may prove valuable over time, we believe that for these classes of tasks, dependable whole-body behavior, compliance, and ease of deployment are more immediate constraints than dexterous manipulation. 

Similar considerations informed decisions around sensing and morphology. We excluded wrist-mounted cameras to reduce system complexity and integration burden. For many applications, head-mounted RGB-D sensing combined with whole-body teleoperation and mapping provides sufficient perceptual coverage for a broad set of research. The robot's height, which is shorter than an adult human, imposes limitations in certain environments, but significantly improves safety during close interaction and reduces the kinetic energy involved in falls or unintended contact. For operation in homes, labs, and public spaces, we believe this trade-off meaningfully shifts the balance toward safe, approachable, and repeatable use. 

Taken together, these design choices address long-standing barriers to broader participation and reliable research in humanoid robotics. By providing a platform that is safe to operate around people, robust enough for day-to-day use, and structured to support reproducible deployment, we seek to broaden access to embodied robotics beyond specialized labs and short-term demonstrations. Lowering these practical and safety-related barriers is, in our view, a prerequisite for rigorous research on HRI, learning from real-world experience, and long-horizon autonomy. By foregrounding safety and expressivity, the platform supports a shift toward sustained, trustworthy humanoid operation in environments designed for regular human interaction.

\addtolength{\textheight}{-0cm}   



\section*{Fauna Team}
Diego Aldarondo,
Ana Pervan,
Daniel Corbalan,
Dave Petrillo,
Bolun Dai,
Aadhithya Iyer,
Nina Mortensen,
Erik Pearson,
Sridhar Pandian Arunachalam,
Emma Reznick,
David Weis,
Jacob Davison,
Samuel Patterson,
Tess Carella,
Michael Suguitan,
David Ye,
Oswaldo Ferro,
Nilesh Suriyarachchi,
Spencer Ling,
Erik Su,
Daniel Giebisch,
Peter Traver,
Sam Fonseca,
Mack Mor,
Rohan Singh,
Sertac Guven,
Kangni Liu,
Yaswanth Kumar Orru,
Ashiq Rahman Anwar Batcha,
Shruthi Ravindranath,
Silky Arora,
Hugo Ponte,
Dez Hernandez,
Utsav Chaudhary,
Zack Walker,
Michael Kelberman,
Ivan Veloz,
Christina Santa Lucia,
Kat Casale,
Helen Han,
Michael Gromis,
Michael Mignatti,
Jason Reisman,
Kelleher Guerin,
Dario Narvaez,
Christopher Anderson,
Anthony Moschella,
Robert Cochran,
Josh Merel.

\section*{Acknowledgements}
Thanks to Keara Inecia, Samantha Bickart, and Ray Cruz for supporting contributions.


{
\hypersetup{urlcolor=black}
\bibliography{bibliography.bib}

@article{shi2025toddlerbot,
  author  = {Haochen Shi and Weizhuo Wang and Shuran Song and C. Karen Liu},
  title   = {Toddler{B}ot: Open-Source {ML}-Compatible Humanoid Platform for Loco-Manipulation},
  journal = {CoRR},
  volume  = {abs/2502.00893},
  year    = {2025}
}

@inproceedings{Grandia2024,
  author    = {Ruben Grandia and Espen Knoop and Michael A. Hopkins and Georg Wiedebach and Jared Bishop and Steven Pickles and David M{\"{u}}ller and Moritz B{\"{a}}cher},
  title     = {Design and Control of a Bipedal Robotic Character},
  booktitle = {Proceedings of Robotics: Science and Systems XX, Delft, The Netherlands},
  month     = {July},
  year      = {2024}
}

@inproceedings{liao2024berkeley,
  author    = {Qiayuan Liao and Bike Zhang and Xuanyu Huang and Xiaoyu Huang and Zhongyu Li and Koushil Sreenath},
  title     = {{Berkeley Humanoid}: {A} Research Platform for Learning-Based Control},
  booktitle = {Proceedings of {IEEE} International Conference on Robotics and Automation, Atlanta, GA, USA},
  pages     = {2897--2904},
  month     = {May},
  year      = {2025}
}

@misc{rudin2022learningwalkminutesusing,
      title={Learning to Walk in Minutes Using Massively Parallel Deep Reinforcement Learning}, 
      author={Nikita Rudin and David Hoeller and Philipp Reist and Marco Hutter},
      year={2022},
      eprint={2109.11978},
      archivePrefix={arXiv},
      primaryClass={cs.RO},
      url={https://arxiv.org/abs/2109.11978}, 
}

@misc{portela2024learningforcecontrollegged,
      title={Learning Force Control for Legged Manipulation}, 
      author={Tifanny Portela and Gabriel B. Margolis and Yandong Ji and Pulkit Agrawal},
      year={2024},
      eprint={2405.01402},
      archivePrefix={arXiv},
      primaryClass={cs.RO},
      url={https://arxiv.org/abs/2405.01402}, 
}

@misc{margolis2025softmimiclearningcompliantwholebody,
      title={SoftMimic: Learning Compliant Whole-body Control from Examples}, 
      author={Gabriel B. Margolis and Michelle Wang and Nolan Fey and Pulkit Agrawal},
      year={2025},
      eprint={2510.17792},
      archivePrefix={arXiv},
      primaryClass={cs.RO},
      url={https://arxiv.org/abs/2510.17792}, 
}

@article{jenelten2024dtc,
  title={Dtc: Deep tracking control},
  author={Jenelten, Fabian and He, Junzhe and Farshidian, Farbod and Hutter, Marco},
  journal={Science Robotics},
  volume={9},
  number={86},
  pages={eadh5401},
  year={2024},
  publisher={American Association for the Advancement of Science}
}

@misc{zhang2026ame2agilegeneralizedlegged,
      title={AME-2: Agile and Generalized Legged Locomotion via Attention-Based Neural Map Encoding}, 
      author={Chong Zhang and Victor Klemm and Fan Yang and Marco Hutter},
      year={2026},
      eprint={2601.08485},
      archivePrefix={arXiv},
      primaryClass={cs.RO},
      url={https://arxiv.org/abs/2601.08485}, 
}

@online{Brooks2025HumanoidsDexterity,
  author = {Brooks, Rodney},
  title  = {Why Today’s Humanoids Won’t Learn Dexterity},
  year   = {2025},
  url    = {https://rodneybrooks.com/why-todays-humanoids-wont-learn-dexterity/}
}

@misc{ze2025twist2scalableportableholistic,
      title={TWIST2: Scalable, Portable, and Holistic Humanoid Data Collection System}, 
      author={Yanjie Ze and Siheng Zhao and Weizhuo Wang and Angjoo Kanazawa and Rocky Duan and Pieter Abbeel and Guanya Shi and Jiajun Wu and C. Karen Liu},
      year={2025},
      eprint={2511.02832},
      archivePrefix={arXiv},
      primaryClass={cs.RO},
      url={https://arxiv.org/abs/2511.02832}, 
}

@misc{bjelonic2025bridginggapsystematicsimtoreal,
      title={Towards bridging the gap: Systematic sim-to-real transfer for diverse legged robots}, 
      author={Filip Bjelonic and Fabian Tischhauser and Marco Hutter},
      year={2025},
      eprint={2509.06342},
      archivePrefix={arXiv},
      primaryClass={cs.RO},
      url={https://arxiv.org/abs/2509.06342}, 
}

@inproceedings{zhu2025artemis,
  author    = {Taoyuanmin Zhu and Min Sung Ahn and Dennis W. Hong},
  title     = {{ARTEMIS:} An Open-Source, Full-Sized Humanoid Robot for Dynamic Locomotion},
  booktitle = {Proceedings of 24th {IEEE-RAS} International Conference on Humanoid Robots, Seoul, Republic of Korea},
  pages     = {269--276},
  month     = {September},
  year      = {2025}
}

@inproceedings{Peng_2018,
   title={Sim-to-Real Transfer of Robotic Control with Dynamics Randomization},
   url={http://dx.doi.org/10.1109/ICRA.2018.8460528},
   DOI={10.1109/icra.2018.8460528},
   booktitle={2018 IEEE International Conference on Robotics and Automation (ICRA)},
   publisher={IEEE},
   author={Peng, Xue Bin and Andrychowicz, Marcin and Zaremba, Wojciech and Abbeel, Pieter},
   year={2018},
   month=may, pages={3803–3810} }

@online{RobotisOP3_eManual,
  author = {ROBOTIS},
  title  = {{ROBOTIS e-Manual}},
  year   = {2025},
  url    = {https://emanual.robotis.com/docs/en/platform/op3/introduction/}
}

@article{labbe2019rtabmap,
  author  = {Labb{\'e}, Mathieu and Michaud, Fran{\c{c}}ois},
  title   = {{RTAB-Map} as an Open-Source {LiDAR} and Visual {SLAM} Library for Large-Scale and Long-Term Online Operation},
  journal = {Journal of Field Robotics},
  volume  = {36},
  number  = {2},
  pages   = {416--446},
  year    = {2019}
}

@inproceedings{brohan2022rt,
  author    = {Brohan, Anthony and Brown, Noah and Carbajal, Justice and Chebotar, Yevgen and Dabis, Joseph and Finn, Chelsea and Gopalakrishnan, Keerthana and Hausman, Karol and Herzog, Alex and Hsu, Jasmine and others},
  title     = {{RT-1:} Robotics Transformer for Real-World Control at Scale},
  booktitle = {Proceedings of Robotics: Science and Systems XIX, Daegu, Republic of Korea},
  month     = {July},
  year      = {2023}
}

@article{team2025gemini,
  author  = {{Gemini Robotics Team} and Abdolmaleki, Abbas and Abeyruwan, Saminda and Ainslie, Joshua and Alayrac, Jean-Baptiste and Arenas, Montserrat Gonzalez and Balakrishna, Ashwin and Batchelor, Nathan and Bewley, Alex and Bingham, Jeff and others},
  title   = {{Gemini Robotics 1.5}: Pushing the Frontier of Generalist Robots with Advanced Embodied Reasoning, Thinking, and Motion Transfer},
  journal = {CoRR},
  volume  = {abs/2510.03342},
  year    = {2025}
}

@article{intelligence2025pi05visionlanguageactionmodelopenworld,
  title   = {$\pi_{0.5}$: a Vision-Language-Action Model with Open-World Generalization},
  author  = {{Physical Intelligence} and Kevin Black and Noah Brown and James Darpinian and Karan Dhabalia and Danny Driess and Adnan Esmail and Michael Equi and Chelsea Finn and Niccolo Fusai and Manuel Y. Galliker and Dibya Ghosh and Lachy Groom and Karol Hausman and Brian Ichter and Szymon Jakubczak and Tim Jones and Liyiming Ke and Devin LeBlanc and Sergey Levine and Adrian Li-Bell and Mohith Mothukuri and Suraj Nair and Karl Pertsch and Allen Z. Ren and Lucy Xiaoyang Shi and Laura Smith and Jost Tobias Springenberg and Kyle Stachowicz and James Tanner and Quan Vuong and Homer Walke and Anna Walling and Haohuan Wang and Lili Yu and Ury Zhilinsky},
  journal = {CoRR},
  volume  = {abs/2504.16054},
  year    = {2025}
}

@misc{generalist2025gen0,
  author       = {{Generalist AI Team}},
  title        = {{GEN-0}: Embodied Foundation Models That Scale with Physical Interaction},
  year         = {2025},
  howpublished = {Online},
  url = {https://generalistai.com/blog/nov-04-2025-GEN-0},
}

@article{nvidia2025gr00tn1openfoundation,
  title   = {{GR00T N1}: An Open Foundation Model for Generalist Humanoid Robots},
  author  = {Johan Bjorck and Fernando Castañeda and Nikita Cherniadev and Xingye Da and Runyu Ding and Linxi "Jim" Fan and Yu Fang and Dieter Fox and Fengyuan Hu and Spencer Huang and Joel Jang and Zhenyu Jiang and Jan Kautz and Kaushil Kundalia and Lawrence Lao and Zhiqi Li and Zongyu Lin and Kevin Lin and Guilin Liu and Edith Llontop and Loic Magne and Ajay Mandlekar and Avnish Narayan and Soroush Nasiriany and Scott Reed and You Liang Tan and Guanzhi Wang and Zu Wang and Jing Wang and Qi Wang and Jiannan Xiang and Yuqi Xie and Yinzhen Xu and Zhenjia Xu and Seonghyeon Ye and Zhiding Yu and Ao Zhang and Hao Zhang and Yizhou Zhao and Ruijie Zheng and Yuke Zhu},
  journal = {CoRR},
  volume  = {abs/2503.14734},
  year    = {2025}
}

@misc{Figure2025Helix,
  author       = {{Figure AI}},
  title        = {{Helix}: A Vision-Language-Action Model for Generalist Humanoid Control},
  year         = {2025},
  howpublished = {Online},
  url          = {https://www.figure.ai/news/helix},
}

@inproceedings{ahn2022icanisay,
  title     = {Do As {I} Can, Not As {I} Say: Grounding Language in Robotic Affordances},
  author    = {Michael Ahn and Anthony Brohan and Noah Brown and Yevgen Chebotar and Omar Cortes and Byron David and Chelsea Finn and Chuyuan Fu and Keerthana Gopalakrishnan and Karol Hausman and Alex Herzog and Daniel Ho and Jasmine Hsu and Julian Ibarz and Brian Ichter and Alex Irpan and Eric Jang and Rosario Jauregui Ruano and Kyle Jeffrey and Sally Jesmonth and Nikhil J Joshi and Ryan Julian and Dmitry Kalashnikov and Yuheng Kuang and Kuang-Huei Lee and Sergey Levine and Yao Lu and Linda Luu and Carolina Parada and Peter Pastor and Jornell Quiambao and Kanishka Rao and Jarek Rettinghouse and Diego Reyes and Pierre Sermanet and Nicolas Sievers and Clayton Tan and Alexander Toshev and Vincent Vanhoucke and Fei Xia and Ted Xiao and Peng Xu and Sichun Xu and Mengyuan Yan and Andy Zeng},
  booktitle = {Proceedings of the Conference on Robot Learning, Auckland, New Zealand},
  volume    = {205},
  pages     = {287--318},
  month     = {December},
  year      = {2022}
}

@inproceedings{liang2023codepolicieslanguagemodel,
  author    = {Jacky Liang and Wenlong Huang and Fei Xia and Peng Xu and Karol Hausman and Brian Ichter and Pete Florence and Andy Zeng},
  title     = {{Code as Policies}: Language Model Programs for Embodied Control},
  booktitle = {Proceedings of {IEEE} International Conference on Robotics and Automation, London, UK},
  pages     = {9493--9500},
  month     = {May},
  year      = {2023}
}

@misc{Shenkar2025MenteebotApproach,
  author       = {Shenkar, Tom and Gur, Shir and Wolf, Lior},
  title        = {{Menteebot AI Approach}},
  howpublished = {Online},
  url = {https://menteebot.com/blog/\#menteebot-ai-approach},
  year         = {2025},
}

@misc{BostonDynamics2025LargeBehaviorModels,
  author       = {{Boston Dynamics} and {Toyota Research Institute}},
  title        = {Large Behavior Models and {A}tlas Find New Footing},
  howpublished = {Online},
  url = {https://bostondynamics.com/blog/large-behavior-models-atlas-find-new-footing/},
  year         = {2025},
}

@article{macenski2022robot,
  title   = {{Robot Operating System 2}: Design, Architecture, and Uses in the Wild},
  author  = {Macenski, Steven and Foote, Tully and Gerkey, Brian and Lalancette, Chris and Woodall, William},
  journal = {Science Robotics},
  volume  = {7},
  number  = {66},
  year    = {2022}
}

@article{mittal2025isaac,
  title   = {{I}saac {L}ab: A {GPU}-Accelerated Simulation Framework for Multi-Modal Robot Learning},
  author  = {Mittal, Mayank and Roth, Pascal and Tigue, James and Richard, Antoine and Zhang, Octi and Du, Peter and Serrano-Mu{\~n}oz, Antonio and Yao, Xinjie and Zurbr{\"u}gg, Ren{\'e} and Rudin, Nikita and others},
  journal = {CoRR},
  volume  = {abs/2511.04831},
  year    = {2025}
}

@inproceedings{todorov2012mujoco,
  title     = {Mu{J}o{C}o: A physics engine for model-based control},
  author    = {Todorov, Emanuel and Erez, Tom and Tassa, Yuval},
  booktitle = {Proceedings of {IEEE/RSJ} International Conference on Intelligent Robots and Systems, Vilamoura, Algarve, Portugal},
  pages     = {5026--5033},
  month     = {October},
  year      = {2012}
}

@software{Genesis,
  author = {{Genesis Authors}},
  title  = {{Genesis}: A Generative and Universal Physics Engine for Robotics and Beyond},
  month  = {December},
  year   = {2024},
  url    = {https://github.com/Genesis-Embodied-AI/Genesis}
}

@article{andrychowicz2018learning,
  author  = {Marcin Andrychowicz and Bowen Baker and Maciek Chociej and Rafal J{\'{o}}zefowicz and Bob McGrew and Jakub Pachocki and Arthur Petron and Matthias Plappert and Glenn Powell and Alex Ray and Jonas Schneider and Szymon Sidor and Josh Tobin and Peter Welinder and Lilian Weng and Wojciech Zaremba},
  title   = {Learning dexterous in-hand manipulation},
  journal = {The International Journal of Robotics Research},
  volume  = {39},
  number  = {1},
  year    = {2020}
}

@article{hwangbo2019learning,
  author  = {Hwangbo, Jemin and Lee, Joonho and Dosovitskiy, Alexey and Bellicoso, Dario and Tsounis, Vassilios and Koltun, Vladlen and Hutter, Marco},
  title   = {Learning Agile and Dynamic Motor Skills for Legged Robots},
  journal = {Science Robotics},
  volume  = {4},
  number  = {26},
  year    = {2019}
}

@article{heess2017emergence,
  title   = {Emergence of locomotion behaviours in rich environments},
  author  = {Heess, Nicolas and Tb, Dhruva and Sriram, Srinivasan and Lemmon, Jay and Merel, Josh and Wayne, Greg and Tassa, Yuval and Erez, Tom and Wang, Ziyu and Eslami, SM and others},
  journal = {CoRR},
  volume  = {abs/1707.02286},
  year    = {2017}
}

@misc{Flexion2025ReflectV0,
  author       = {{Flexion Team}},
  title        = {{Flexion Reflect v0 -- Towards Generalizable Robot Autonomy}},
  howpublished = {Online},
  url = {https://flexion.ai/news/flexion-reflect-v0-towards-generalizable-robot-autonomy},
  year         = {2025},
}

@inproceedings{dolgov2008practical,
  title     = {Practical Search Techniques in Path Planning for Autonomous Driving},
  author    = {Dolgov, Dmitri and Thrun, Sebastian and Montemerlo, Michael and Diebel, James},
  booktitle = {Proceedings of the First International Symposium on Search Techniques in Artificial Intelligence and Robotics (STAIR-08), Chicago, IL, USA},
  month     = {June},
  year      = {2008}
}

@article{MacenskiS0G23,
  author  = {Steve Macenski and Shrijit Singh and Francisco Mart{\'{\i}}n and Jonatan Gin{\'{e}}s},
  title   = {Regulated pure pursuit for robot path tracking},
  journal = {Autonomous Robots},
  volume  = {47},
  number  = {6},
  pages   = {685--694},
  year    = {2023}
}

@article{HornungWBSB13,
  author  = {Armin Hornung and Kai M. Wurm and Maren Bennewitz and Cyrill Stachniss and Wolfram Burgard},
  title   = {{OctoMap}: an efficient probabilistic {3D} mapping framework based on octrees},
  journal = {Autonomous Robots},
  volume  = {34},
  number  = {3},
  pages   = {189--206},
  year    = {2013}
}

@article{camurri2020pronto,
  title   = {Pronto: A Multi-Sensor State Estimator for Legged Robots in Real-World Scenarios},
  author  = {Camurri, Marco and Ramezani, Milad and Nobili, Simona and Fallon, Maurice},
  journal = {Frontiers in Robotics and AI},
  year    = {2020},
  volume  = {7},
  pages   = {68}
}

@inproceedings{yang2023cerberus,
  author    = {Yang, Shuo and Zhang, Zixin and Fu, Zhengyu and Manchester, Zachary},
  title     = {Cerberus: Low-Drift Visual-Inertial-Leg Odometry For Agile Locomotion},
  booktitle = {Proceedings of {IEEE} International Conference on Robotics and Automation, London, UK},
  pages     = {4193--4199},
  month     = {May},
  year      = {2023}
}

@article{campos2021orbslam,
  author  = {Campos, Carlos and Elvira, Richard and Rodriguez, Juan J. Gomez and M. Montiel, Jose M. and D. Tardos, Juan},
  title   = {{ORB-SLAM3}: An Accurate Open-Source Library for Visual, Visual–Inertial, and Multimap {SLAM}},
  journal = {IEEE Transactions on Robotics},
  volume  = {37},
  number  = {6},
  pages   = {1874--1890},
  year    = {2021},
  month   = {December}
}

@article{zhang20242dliwslam,
  title   = {{2DLIW-SLAM}:{2D} {LiDAR}-Inertial-Wheel Odometry with Real-Time Loop Closure},
  author  = {Bin Zhang and Zexin Peng and Bi Zeng and Junjie Lu},
  journal = {CoRR},
  volume  = {abs/2404.07644},
  year    = {2024}
}

@inproceedings{xiao2025geoflowslam,
  author    = {Tingyang Xiao and Xiaolin Zhou and Liu Liu and Wei Sui and Wei Feng and Jiaxiong Qiu and Xinjie Wang and Zhizhong Su},
  title     = {{G}eo{F}low-{SLAM}: {A} Robust Tightly-Coupled RGBD-Inertial and Legged Odometry Fusion {SLAM} for Dynamic Legged Robotics},
  booktitle = {Proceedings of {IEEE/RSJ} International Conference on Intelligent Robots and Systems, Hangzhou, China},
  pages     = {15181--15188},
  month     = {October},
  year      = {2025}
}

@inproceedings{yan2024survey,
  title     = {A Survey of {SLAM} based on Submap Strategies},
  author    = {Han Yan},
  booktitle = {Proceedings of the 2024 International Conference on Artificial Intelligence and Communication, Davos, Switzerland},
  year      = {2024},
  month     = {June},
  pages     = {122--131}
}

@article{dellaert2012factor,
  title   = {Factor Graphs and {GTSAM}: A Hands-on Introduction},
  author  = {Dellaert, Frank},
  journal = {Georgia Institute of Technology},
  year    = {2012},
  note    = {{Technical Report GT-RIM-CP\&R-2012-002}},
  url     = {https://arxiv.org/abs/1206.0526}
}

@software{MCP,
  author = {{Anthropic}},
  title  = {{Model Context Protocol}},
  year   = {2024},
  url    = {https://www.anthropic.com/news/model-context-protocol}
}

@software{openWakeWord,
  author = {{openWakeWord Team}},
  title  = {{openWakeWord: An open-source audio wake word (or phrase) detection framework with a focus on performance and simplicity.}},
  year   = {2023},
  url    = {https://github.com/dscripka/openWakeWord}
}

@software{DeepgramFlux,
  author = {{Deepgram Team}},
  title  = {{Introducing Flux: Conversational Speech Recognition to Solve the Biggest Problem in Voice Agents-Interruptions}},
  year   = {2025},
  url    = {https://deepgram.com/learn/introducing-flux-conversational-speech-recognition}
}

@software{ElevenLabsV2,
  author = {{ElevenLabs Team}},
  title  = {{Eleven Multilingual v2}},
  year   = {2025},
  url    = {https://elevenlabs.io/blog/eleven-multilingual-v2}
}

@software{NVIDIARiva,
  author = {{NVIDIA}},
  title  = {{NVIDIA Riva: A collection of GPU-accelerated microservices for building real-time, customizable speech AI applications}},
  year   = {2025},
  url    = {http://nvidia.com/en-us/ai-data-science/products/riva/}
}

@article{mueller2025olaf,
  title   = {Olaf: Bringing an Animated Character to Life in the Physical World},
  author  = {M{\"u}ller, David and Knoop, Espen and Mylonopoulos, Dario and Serifi, Agon and Hopkins, Michael A. and Grandia, Ruben and B{\"a}cher, Moritz},
  journal = {CoRR},
  volume  = {abs/2512.16705},
  year    = {2025}
}

@misc{rmwzenoh,
  author       = {{ROS 2 Community}},
  title        = {{RMW Zenoh}: A Zenoh-based ROS 2 middleware implementation},
  year         = {2024},
  url          = {https://github.com/ros2/rmw_zenoh},
}

@software{cbor,
  author = {{IETF}},
  title  = {{Concise Binary Object Representation (CBOR)}},
  year   = {2020},
  url    = {https://cbor.io/}
}

@software{foxglove,
  author = {{Foxglove Technologies}},
  title  = {{Foxglove: Visualization and debugging for robotics}},
  year   = {2024},
  url    = {https://foxglove.dev/}
}

@software{pink2024,
  title   = {{Pink: Python inverse kinematics based on Pinocchio}},
  author  = {Caron, Stéphane and De Mont-Marin, Yann and Budhiraja, Rohan and Bang, Seung Hyeon and Domrachev, Ivan and Nedelchev, Simeon and Du, Peter and Escande, Adrien and Vaillant, Joris and Wingo, Bruce},
  license = {Apache-2.0},
  url     = {https://github.com/stephane-caron/pink},
  version = {3.5.0},
  year    = {2025}
}

@article{intelligence2025pistar06,
  title   = {$\pi^{*}_{0.6}$: a {VLA} That Learns From Experience},
  author  = {{Physical Intelligence} and Kevin Black and Noah Brown and James Darpinian and Karan Dhabalia and Danny Driess and Adnan Esmail and Michael Equi and Chelsea Finn and Niccolo Fusai and Manuel Y. Galliker and Dibya Ghosh and Lachy Groom and Karol Hausman and Brian Ichter and Szymon Jakubczak and Tim Jones and Liyiming Ke and Devin LeBlanc and Sergey Levine and Adrian Li-Bell and Mohith Mothukuri and Suraj Nair and Karl Pertsch and Allen Z. Ren and Lucy Xiaoyang Shi and Laura Smith and Jost Tobias Springenberg and Kyle Stachowicz and James Tanner and Quan Vuong and Homer Walke and Anna Walling and Haohuan Wang and Lili Yu and Ury Zhilinsky},
  journal = {CoRR},
  volume  = {abs/2511.14759},
  year    = {2025}
}

@inproceedings{ross2011dagger,
  title     = {A Reduction of Imitation Learning and Structured Prediction to No-Regret Online Learning},
  author    = {Ross, St{\'e}phane and Gordon, Geoffrey and Bagnell, Drew},
  booktitle = {Proceedings of the Fourteenth International Conference on Artificial Intelligence and Statistics, Fort Lauderdale, FL, USA},
  pages     = {627--635},
  year      = {2011},
  month     = {April},
  volume    = {15}
}

@inproceedings{iyer2024openteach,
  title     = {{OPEN TEACH}: {A} Versatile Teleoperation System for Robotic Manipulation},
  author    = {Iyer, Aadhithya and Peng, Zhuoran and Dai, Yinlong and Guzey, Irmak and Haldar, Siddhant and Chintala, Soumith and Pinto, Lerrel},
  booktitle = {Proceedings of the Conference on Robot Learning, Munich, Germany},
  volume    = {270},
  pages     = {2372--2395},
  month     = {November},
  year      = {2024}
}

@inproceedings{li2025amo,
  title     = {{AMO}: Adaptive Motion Optimization for Hyper-Dexterous Humanoid Whole-Body Control},
  author    = {Li, Jialong and Cheng, Xuxin and Huang, Tianshu and Yang, Shiqi and Qiu, Rizhao and Wang, Xiaolong},
  booktitle = {Proceedings of Robotics: Science and Systems XXI, Los Angeles, CA, USA},
  month     = {June},
  year      = {2025}
}

@inproceedings{jakobi1995noise,
  title={Noise and the reality gap: The use of simulation in evolutionary robotics},
  author={Jakobi, Nick and Husbands, Phil and Harvey, Inman},
  booktitle={European conference on artificial life},
  pages={704--720},
  year={1995},
  organization={Springer}
}
}
\bibliographystyle{IEEEtran}

\clearpage

\section*{Supplement}

\begin{figure}[h!]
  \centering
  \includegraphics[width=\linewidth]{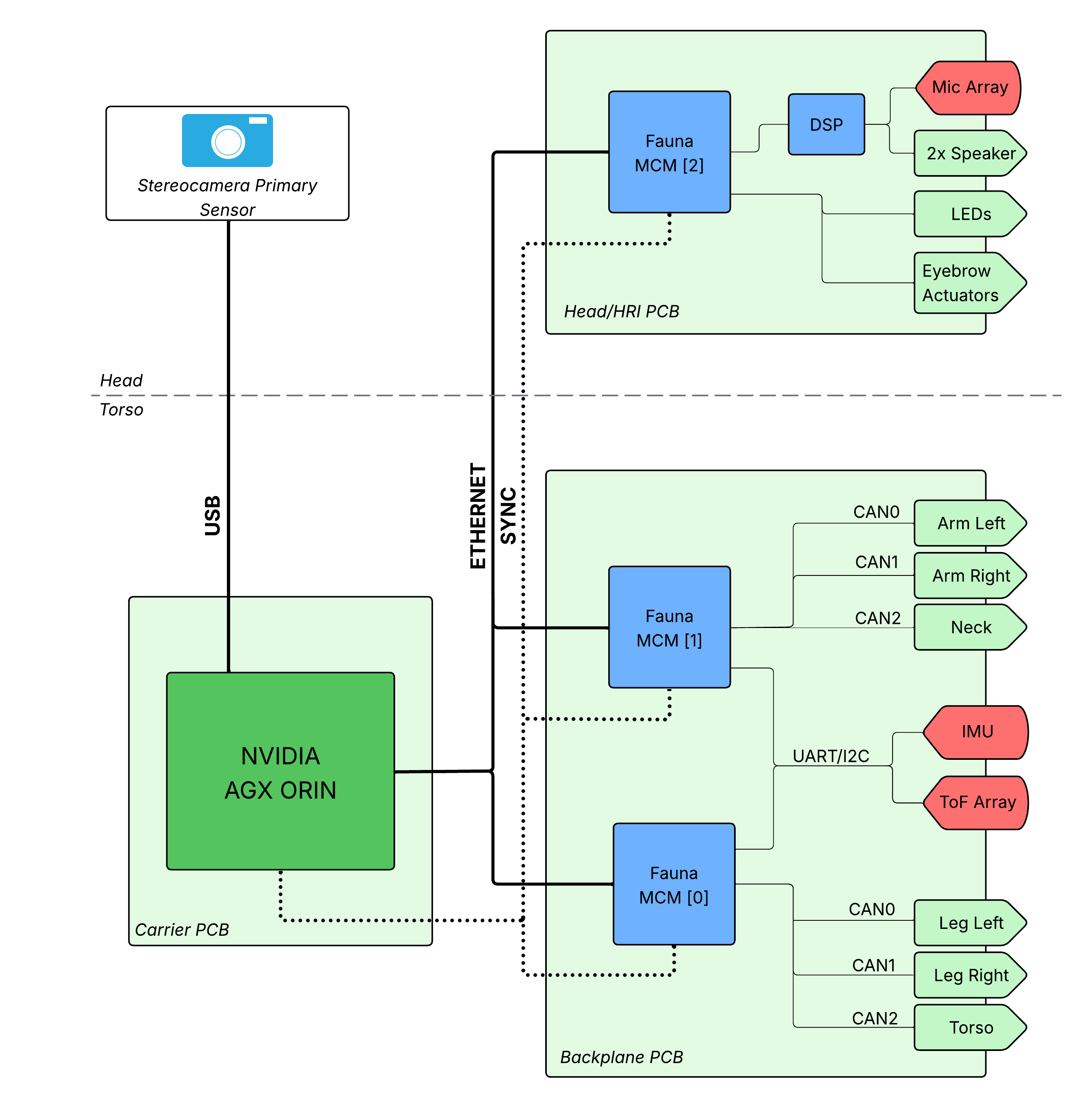}
  \caption{\textbf{Compute architecture.} The AGX communicates with custom motor control modules (MCMs), microcontrollers, which in turn communicate with motors and sensors. The MCMs mount to a custom backplane, or ``motherboard'', which also regulates power to the motors and computers.}
  \label{fig:compute}
\end{figure}

\begin{table}[h!]
\centering
\caption{\textbf{Technical Specifications.}}
\label{table:tech_specs}
\begin{tabular}{|l|p{5cm}|}
\hline
\textbf{Category} & \textbf{Details} \\ \hline

\textbf{Dimensions} & Standing Height: \SI{107}{cm} \\
 & Width: \SI{45}{cm} \\
 & Depth: \SI{21}{cm} \\ \hline

\textbf{Mass} & \SI{22.7}{kg} \\ \hline

\textbf{Compute} & NVIDIA Jetson AGX Orin 64GB \\ \hline

\textbf{Sensing} & ZED2i RGB-D stereo camera \\ & 4 × VL53L8CX ToF \\ & 9-axis IMU \\ & 4 × MEMS microphone array \\ \hline

\textbf{Actuation} & 29 degrees of freedom, \\ & including (2x) eyebrows \\ \hline

\textbf{Power / Battery} & \SI{46.8}{V} Nominal DC system \\ & \SI{5000}{mAh} (standard) \\ & or \SI{10000}{mAh} (extended run) \\ & Li-ion battery (Molicel P50B cells) \\ & Runtime 3-3.5 h \\ \hline

\textbf{Safety} & E-Stop \\ \hline

\textbf{Environmental} & Operating Temperature: \SIrange{10}{30}{\degreeCelsius} \\
 & Storage Temperature: \SIrange{5}{35}{\degreeCelsius} \\
 & Humidity: 10\%--90\% RH noncondensing \\
 & Indoor use only \\ \hline

\end{tabular}
\end{table}

\begin{figure}[!t]
  \centering
  \includegraphics[width=\linewidth]{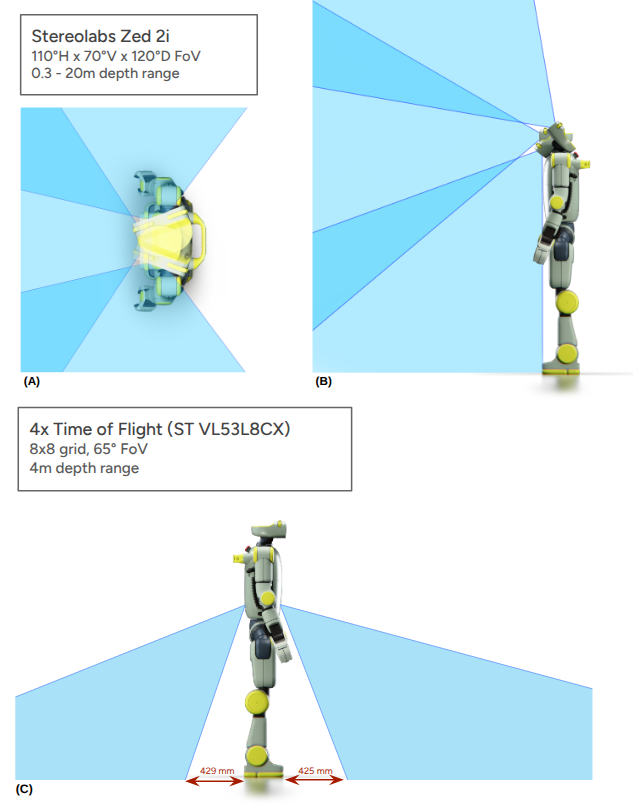}
  \caption{\textbf{Field-of-view.} (A) top-down view of the primary sensor horizontal FoV, for 3 representative neck yaw angles (B) profile view of the primary sensor vertical FoV, for 3 representative neck pitch angles (C) profile view of the torso time-of-flight obstacle sensor FoV}
  \label{fig:fov}
\end{figure}

\FloatBarrier

\begin{table}[!h]
\centering
\caption{\textbf{Payloads.} Per-arm theoretical maximum payloads; actual supported payloads may differ depending on the teleoperation controller.}
\label{table:payloads}
\begin{tabular}{|l|c|c|c|}
\hline
\textbf{Configuration} & \textbf{Max (kg)} & \textbf{Nominal (kg)} & \textbf{Rated (kg)} \\
 & \textbf{($<$10s)} & \textbf{($<$90s)} & \textbf{(indefinitely)} \\ \hline
Forward Raise & 3.7 & 1.75 & 0.7 \\ \hline
Lateral Raise & 5.6 & 2.5 & 0.0\textsuperscript{*} \\ \hline
Bent Elbow & 11.2 & 5.0 & 1.1 \\ \hline
\end{tabular}
\begin{flushleft}
\textit{\textsuperscript{*} It is not recommended to maintain the Lateral Raise configuration indefinitely, with or without payload.}
\end{flushleft}
\vspace{-0.5em}
\end{table}

\end{document}